\documentclass{ieeeaccess}
\usepackage{cite}
\usepackage{amsmath,amssymb,amsfonts}
\usepackage{algorithmic}
\usepackage{graphicx}
\usepackage{textcomp}
\usepackage{csquotes}
\usepackage{multicol}
\usepackage[export]{adjustbox}

\usepackage{placeins}
\usepackage{algorithm}
\usepackage{multirow}
\usepackage{array}
\usepackage{bibentry}

\usepackage{bm}
\makeatletter
\AtBeginDocument{\DeclareMathVersion{bold}
\SetSymbolFont{operators}{bold}{T1}{times}{b}{n}
\SetSymbolFont{NewLetters}{bold}{T1}{times}{b}{it}
\SetMathAlphabet{\mathrm}{bold}{T1}{times}{b}{n}
\SetMathAlphabet{\mathit}{bold}{T1}{times}{b}{it}
\SetMathAlphabet{\mathbf}{bold}{T1}{times}{b}{n}
\SetMathAlphabet{\mathtt}{bold}{OT1}{pcr}{b}{n}
\SetSymbolFont{symbols}{bold}{OMS}{cmsy}{b}{n}
\renewcommand\boldmath{\@nomath\boldmath\mathversion{bold}}}
\makeatother

\def\BibTeX{{\rm B\kern-.05em{\sc i\kern-.025em b}\kern-.08em
    T\kern-.1667em\lower.7ex\hbox{E}\kern-.125emX}}

\begin{document}

\title{PersonalAI: A Systematic Comparison of Knowledge Graph Storage and Retrieval Approaches for Personalized LLM Agents}
\author{\uppercase{Mikhail Menschikov}\authorrefmark{1}, \uppercase{Dmitry Evseev}\authorrefmark{1}, Victoria Dochkina\authorrefmark{2}, Ruslan Kostoev\authorrefmark{2}, Ilia Perepechkin\authorrefmark{2}, Petr Anokhin\authorrefmark{3}, Nikita Semenov \authorrefmark{1} and Evgeny Burnaev\authorrefmark{1,3}}

\address[1]{Skoltech, Moscow, Russia}
\address[2]{Public Joint Stock Company ``Sberbank of Russia'', Moscow, Russia}
\address[3]{AIRI, Moscow, Russia}


\corresp{Corresponding author: Mikhail Menschikov (e-mail: m.menschikov@skoltech.ru).}

\tfootnote{The work was supported by the grant for research centers in the field of AI provided by the Ministry of Economic Development of the Russian Federation in accordance with agreement 000000C313925P4F0002 and the agreement with Skoltech No.~139-10-2025-033.}

\begin{abstract}
Personalizing language models by effectively incorporating user interaction history remains a central challenge in the development of adaptive AI systems. While large language models (LLMs), combined with Retrieval-Augmented Generation (RAG), have improved factual accuracy, they often lack structured memory and fail to scale in complex, long-term interactions. To address this, we propose a flexible external memory framework based on a knowledge graph that is constructed and updated automatically by the LLM. Building upon the AriGraph architecture, we introduce a novel hybrid graph design that supports both standard edges and two types of hyper-edges, enabling rich and dynamic semantic and temporal representations. Our framework also supports diverse retrieval mechanisms, including A*, WaterCircles traversal, beam search, and hybrid methods, making it adaptable to different datasets and LLM capacities. We evaluate our system on TriviaQA, HotpotQA, DiaASQ benchmarks and demonstrate that different memory and retrieval configurations yield optimal performance depending on the task. Additionally, we extend the DiaASQ benchmark with temporal annotations and internally contradictory statements, showing that our system remains robust and effective in managing temporal dependencies and context-aware reasoning.
\end{abstract}


\titlepgskip=-21pt

\maketitle

\section{Introduction}
\PARstart{R}{ecent} advances in large language models (LLMs) have sparked growing interest in personalized AI systems capable of adapting to users based on their interaction history. Central to personalization is the challenge of encoding, storing, and retrieving relevant information over long time horizons in a manner that supports efficient reasoning and response generation. While Retrieval-Augmented Generation (RAG) has become a widely used solution, enhancing factual recall by appending retrieved content to prompts, it remains limited by its unstructured nature and weak support for semantic relationships across stored memories.

In this work, we introduce a flexible graph-based memory framework designed to overcome these limitations by enabling structured, customizable representations of long-term memory and supporting advanced reasoning capabilities. Unlike traditional RAG pipelines that rely on dense vector similarity over raw text chunks, our system supports multiple memory formats: nodes, knowledge triples, thesis statements, episodic traces and dynamically organizes them into a knowledge graph. This structure allows the agent to represent, update and access semantic and temporal relationships with far greater control and interpretability.

Equally important, our framework supports a pluggable retrieval interface with multiple traversal mechanisms, including three variations of A* search, WaterCircles traversal, BeamSearch, and hybrid strategies to adapt retrieval behavior to task demands and model capacity. We demonstrate that different memory and retrieval configurations yield optimal performance on different benchmarks and LLM scales, highlighting the versatility of our approach.

We build our system on top of the AriGraph architecture \cite{anokhin2024arigraphlearningknowledgegraph}, originally developed for LLM agents in interactive text environments such as TextWorld. AriGraph continuously maintains a structured knowledge base by extracting triples from observations, pruning outdated or redundant facts, and integrating episodic and semantic vertices into a unified graph. Our framework extends this foundation by allowing customization of both memory construction (memorization) and search (retrieval) modules, supporting task-specific tuning and component evaluation.

In summary, our main contributions are as follows:
\begin{enumerate}
\item We present a highly flexible external memory architecture that can be tuned via orthogonal hyperparameters for storage and retrieval.
\item We propose and evaluate six retrieval methods over a structured knowledge graph, achieving superior performance across various datasets compared to GraphRAG baselines.
\item We enhance the DiaASQ benchmark by incorporating temporal structures into dialogue representations and demonstrate that our framework can leverage such structures to improve temporal reasoning.
\end{enumerate}

This work provides a general and extensible framework for integrating long-term memory and adaptive reasoning into LLM agents, advancing the state of personalized and context-aware language generation.

\section{Related Work}

In recent years, substantial advancements have been made in open-domain question answering (QA) and the personalization of language models. Techniques leveraging Wikipedia as a broad knowledge source~\cite{chen-etal-2017-reading} have successfully incorporated large-scale machine reading, integrating document retrieval with textual comprehension. Furthermore, dense representation-based methods for passage retrieval~\cite{karpukhin-etal-2020-dense} have demonstrated superior efficacy compared to traditional sparse retrieval approaches, such as TF-IDF and BM25, particularly in scenarios with sufficient training data.

The development of dense retrievers has demonstrated significant progress, particularly through integration of contrastive learning with unsupervised settings, which has shown promising results across diverse scenarios and outperformed conventional approaches such as BM25~\cite{robertson2009probabilistic} and Contriever~\cite{izacard2021contriever}. Concurrently, pre-trained language models incorporating non-parametric memory access have been proposed for knowledge-intensive tasks. A notable example is retrieval-augmented generation (RAG) models, which integrate parametric and non-parametric memory mechanisms to improve question-answering performance~\cite{lewis2021retrievalaugmentedgenerationknowledgeintensivenlp}.

Recent advancements in unsupervised dense retrieval models, such as ART, have demonstrated the ability to achieve state-of-the-art performance while eliminating reliance on labeled training data~\cite{sachan-etal-2023-questions}. In the domain of knowledge graph-based approaches, frameworks such as GraphReader~\cite{li2024graphreaderbuildinggraphbasedagent} incorporate structured reasoning mechanisms to facilitate knowledge extraction and representation, with a particular emphasis on enhancing long-context reasoning capabilities.

For personalized models, AriGraph~\cite{anokhin2024arigraphlearningknowledgegraph} introduces a framework that integrates episodic memory and long-term planning using knowledge graphs. Similarly, HippoRAG~\cite{gutiérrez2024hipporagneurobiologicallyinspiredlongterm} employs personalized algorithms to improve question-answering (QA) performance by constructing semantic graphs, demonstrating notable advancements over conventional extraction methods. MemWalker~\cite{chen2023walkingmemorymazecontext} and RAPTOR~\cite{sarthi2024raptorrecursiveabstractiveprocessing} address challenges associated with context length, proposing architectures capable of efficiently traversing and consolidating information from large-scale documents.

Additionally, ReadAgent~\cite{lee2024humaninspiredreadingagentgist} addresses the challenge of processing long-text contexts by structuring content into memory episodes. Meanwhile, KGP~\cite{wang2023knowledgegraphpromptingmultidocument} proposes Knowledge Graph Prompting, a method that enhances multi-document question answering by constructing knowledge graphs to improve contextual reasoning. 

These advancements demonstrate a sustained emphasis on improving language models' comprehension, retrieval, and personalization capabilities. Such progress facilitates the development of more sophisticated systems that leverage knowledge graphs to deliver personalized and contextually enriched interactions.

\section{Methods}

\subsection{Memory Structure}

\begin{figure*}[th!]
\centering
\includegraphics[width=0.85\linewidth]{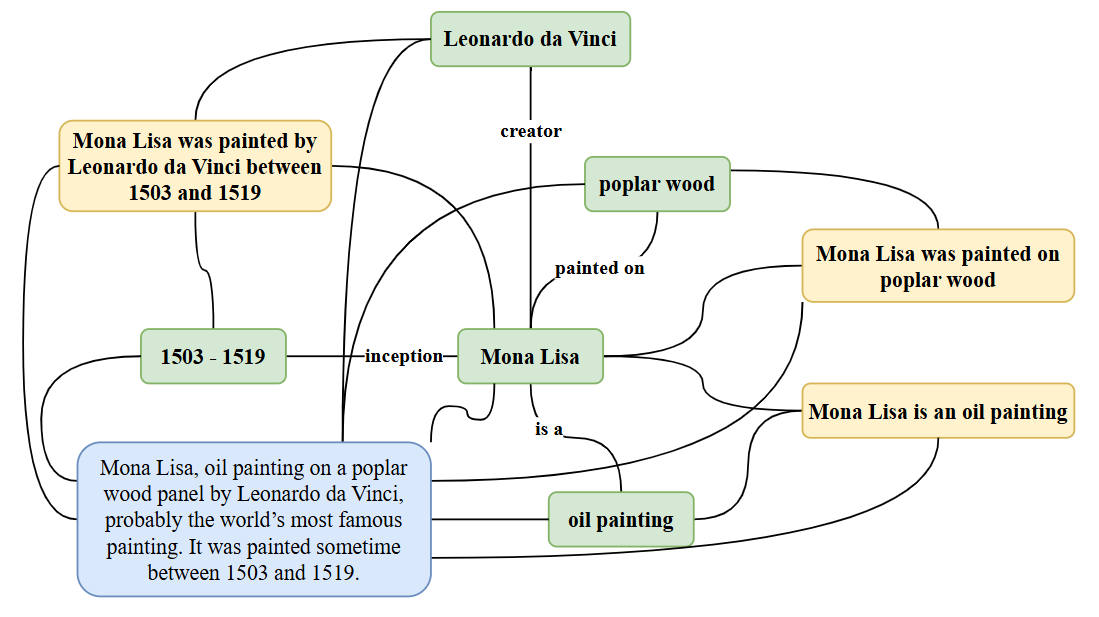}
\caption{Example of a graph fragment, constructed from natural language text using our method, with object (green), thesis (yellow) and episodic (blue) vertices}
\label{fig:graph_example}
\end{figure*}

In this study, we employ a graph knowledge base as external memory to enhance a large language model’s (LLM) question-answering capabilities. The memory model $G = (V_o, E_o, V_t, E_t, V_e, E_e)$ consists of \textit{semantic} $(V_o, E_o, V_t, E_t)$ and \textit{episodic} $(V_e, E_e)$ memory vertices and edges. In turn, semantic vertices and edges are classified into \textit{theses} and \textit{objects}. To capture and structure information from weakly structured natural language texts $d_i$, this memory is constructed automatically by the LLM. This memory graph provides a comprehensive representation of the original texts and comprises the following elements (see Figure \ref{fig:graph_example}):
\begin{enumerate}
    \item $V_o$ is a set of object vertices. Each object vertex represents an atomic concept extracted from the corresponding $d_i$.
    \item $E_o$ is a set of object edges. An object edge is a tuple $(v, rel, u)$, where $v$ and $u$ are object vertices and $rel$ is a text-attributed relationship between them that captures their direct association. Object edges essentially represent triples integrated into the \textit{semantic memory}.
    \item $V_t$ is a set of thesis vertices. Each thesis vertex encapsulates a complete atomic thought expressed in the corresponding $d_i$.
    \item $E_t$ is a set of thesis edges. Thesis edges serve as hyper-edges, linking the corresponding set of object vertices extracted from the same $d_i$ and belonging to the given $v^j_t$.
    \item $V_e$ is a set of episodic vertices. Each episodic vertex corresponds to an original text passage ($v^i_e = d_i$) and serves as a hyper-edge linking related vertices.
    \item $E_e$ is a set of episodic edges. Each episodic edge $e^i_e = (v^i_e, V^i_s)$ connects all semantic vertices $V^i_s$ extracted from $d_i$ through the corresponding episodic vertex $v^i_e$.
\end{enumerate}

\subsection{Memory Construction}

\begin{figure*}[th!]
	\centering
	\includegraphics[width=.85\textwidth]{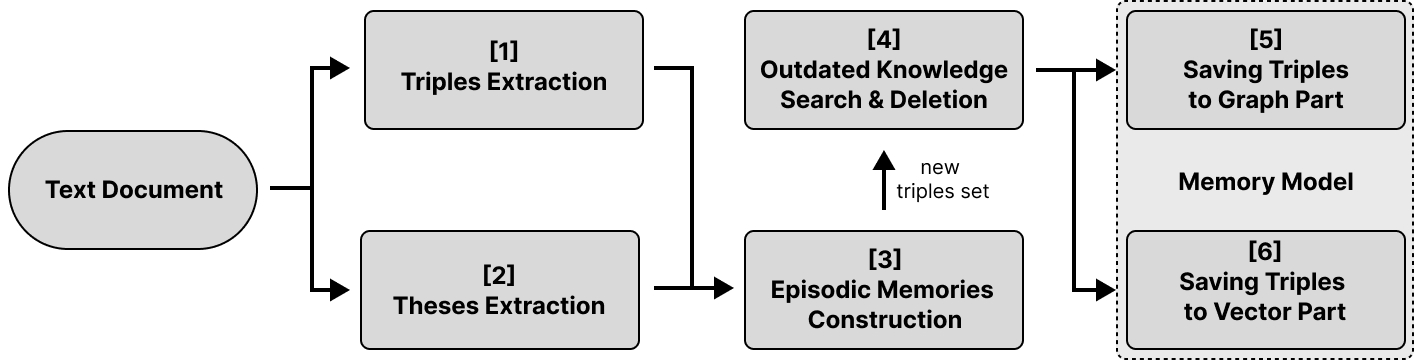}
	\caption{High level architecture of proposed Memorize pipeline for LLM-based triples extraction from unstructured texts on natural language and memory construction} 				
	\label{fig:memipeline_diagram}
\end{figure*}

Using the terminology defined above, the process of constructing a memory graph from weakly structured sources can be decomposed into three key steps: (1) formulating vertices and their edges, (2) generating hyper-edges using an LLM, and (3) parsing the LLM’s output to store extracted information in a structured format (e.g., as subject-relation-object triples). The LLM prompts used to extract semantic memories from textual sources are provided in Appendix~\ref{app:mem_add_prompts}. The full memory construction pipeline (Memorize pipeline) is shown in Figure \ref{fig:memipeline_diagram}.

The identification and retrieval of outdated information within the memory are performed through the following procedure. First, the vertices present in the newly extracted triples are compared against existing vertices in the memory graph to detect matches. Upon identifying matching vertices, these serve as the initial set for a breadth-first search (BFS), which traverses all associated standard and hyper-edges. Subsequently, a specialized prompt is employed to instruct the LLM to update the retrieved knowledge with newly extracted data (see Appendix~\ref{app:mem_upd_prompts} for the corresponding LLM prompts). If any triples are successfully updated, the corresponding outdated instances are removed from the memory model.

\subsection{Information Search in Memory}

\begin{figure*}[th!]
	\centering
	\includegraphics[width=.85\textwidth]{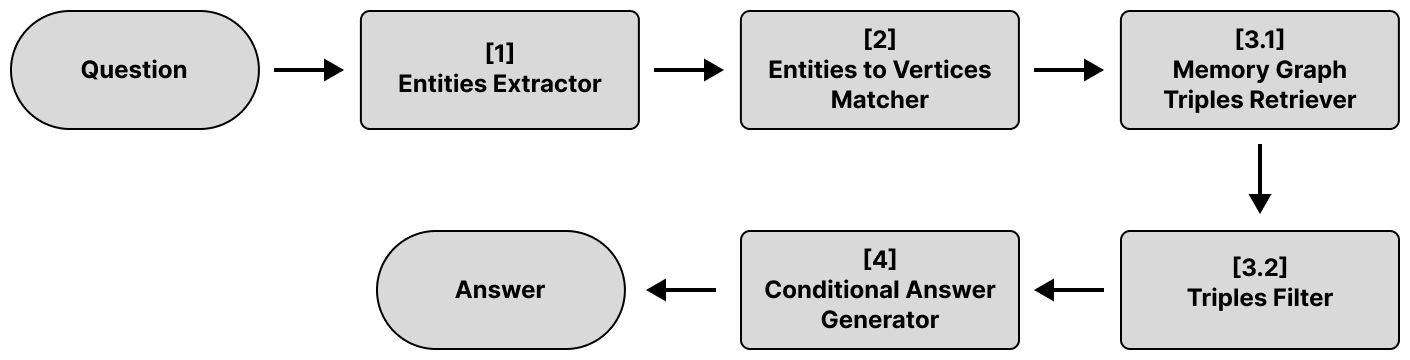}
	\caption{High level architecture of proposed QA pipeline for generating answers to the questions based on constructed memory graph} 				
	\label{fig:weak_qapipeline_diagram}
\end{figure*}

The information search pipeline (QA pipeline) for the memory model is shown in Figure \ref{fig:weak_qapipeline_diagram}.

As illustrated in Figure \ref{fig:weak_qapipeline_diagram}, the pipeline comprises four primary stages, with the third stage further subdivided into two substages. The QA pipeline begins by accepting a natural language question as input, which is then processed by the \textbf{Entities Extractor} module to extract key entities. In the second stage, these entities are passed to the \textbf{Entities to Vertices Matcher} module, where they are aligned with corresponding vertices in the memory graph.

The third stage involves two sequential operations: retrieval and filtering of graph triples based on semantic similarity to the input question. First, the \textbf{Memory Graph Triples Retriever} module initiates a graph traversal algorithm, using the matched entities from the second stage as starting vertices to retrieve a set of candidate triples. Subsequently, the \textbf{Triples Filter} module ranks these triples by computing their semantic similarity to the question via vector embeddings, retaining only the top $N$ (a predefined hyperparameter) most relevant items. 

Finally, in the fourth stage, the \textbf{Conditional Answer Generator} module synthesizes a natural language answer conditioned on the retrieved and filtered triples. The output of the QA pipeline is the generated answer in string format. The LLM prompts used by these modules are provided in Appendix~\ref{app:weak_qa_prompts}.

This pipeline architecture is designed based on three key considerations. First, to achieve an accurate initial approximation of the relevant subgraph, it is essential to align the key entities in the input question with semantically similar vertices in the existing memory graph. The reasoning behind this is that the information necessary for generating a correct response is typically localized within the subgraph containing these key entities. Second, the triples extracted from the knowledge graph exhibit only weak conditioning on the input question, limiting their direct applicability. Third, large language models (LLMs) are constrained by a fixed maximum sequence length for processing inputs in a single inference step, necessitating efficient retrieval and subgraph selection strategies.

\subsection{Retrieval Algorithms}
The primary role of the constructed memory graph in our proposed system is to enable the accurate retrieval of information required for responding to specific user questions. This process necessitates a careful optimization between two key criteria: relevance and completeness. Relevance ensures that all retrieved information is directly pertinent to the question, whereas completeness guarantees the inclusion of all necessary contextual data, even if some retrieved elements may be extraneous. Consequently, striking an optimal balance between these factors is essential for achieving efficient and effective knowledge extraction. 

To accomplish this, we design and implement advanced retrieval algorithms capable of dynamically balancing the trade-off between completeness and relevance through configurable parameter settings. These algorithms constitute a critical component of the question-answering pipeline, systematically traversing the memory graph to aggregate information essential for generating precise and contextually appropriate responses.

\textbf{A*.} The A* algorithm is a widely used method for graph traversal, particularly valued for its ability to efficiently identify shortest paths between vertices in a graph. In the context of our question-answering pipeline, this algorithm extracts and retains triples encountered along these shortest paths while eliminating duplicates based on content. We hypothesize that the triples obtained through this process contain the information necessary for generating accurate responses. For traversal, we treat our memory graph as unweighted and undirected, using a constant distance metric between adjacent vertices. To optimize pathfinding efficiency, we evaluate three distinct heuristics for the $h$-metric:
\begin{enumerate}
    \item \textbf{Inner Product (IP)}: This heuristic computes the dot product between the embeddings of the current and target vertices.
    \item \textbf{Weighted Shortest Path:} This approach scales the inner product metric by the length of the shortest path, determined via Breadth-First Search (BFS).
    \item \textbf{Averaged Weighted Shortest Path:} This heuristic calculates the average of inner product distances between adjacent vertices along the path from the start vertex to the current vertex, as well as the direct path from the current vertex to the target vertex, and further weights this value by the BFS-derived shortest path length.
\end{enumerate}

\textbf{WaterCircles.} This graph-based extraction method employs a breadth-first search (BFS) algorithm to retrieve relevant knowledge. Query entities are first mapped to their corresponding vertices in the memory graph. The algorithm begins by exploring vertices adjacent to the initial vertices and iteratively expands to neighboring vertices in subsequent steps, constructing outward-radiating paths. When paths originating from different starting vertices intersect, the triples formed at these intersections are aggregated into a primary list, while all traversed triples are compiled into a secondary list. The algorithm ultimately returns a subset of triples, selecting $N$ from the primary list and $K$ from the secondary list, where $N$ and $K$ are configurable hyperparameters.

When the memory graph incorporates not only direct object relations (\textit{object} triples), but also associations between objects and text fragments (represented as \textit{thesis} and \textit{episodic} triples), a modified breadth-first search (BFS) algorithm is employed as follows:
\begin{enumerate}
    \item \textbf{Traversal Initialization}: The search begins at vertices that match entities from the input question.
    \item \textbf{Text Fragment Analysis}: During traversal, identified text fragments are examined for occurrences of other question entities, distinct from the entity that originated the path to the given fragment. For each fragment, the number of detected entities, denoted as $N_{intersections}$, is computed.
    \item \textbf{Ranking Triplets}: The list of \textit{thesis} and \textit{episodic} triplets is then sorted in descending order based on $N_{intersections}$.
\end{enumerate}

This triplet extraction strategy aims to enhance relevance and accuracy in retrieving information from the memory graph, thereby bolstering the effectiveness of AI-driven question-answering systems.

\textbf{BeamSearch.} Given a starting vertex, the algorithm constructs $N$ (a hyperparameter) semantically relevant paths in response to the input question. This approach is inspired by beam search, a token generation strategy commonly employed in large language model (LLM) inference. The resulting paths are consolidated into a single list, with duplicate triples removed based on their string content. The traversal process is governed by the following hyperparameters:
\begin{enumerate}
\item \textbf{Max depth}: The maximum allowable depth for path construction.  
\item \textbf{Max paths}: The maximum number of paths to generate.  
\item \textbf{Same Path Intersection by Vertex}: If enabled, a path may revisit a vertex; otherwise, vertex revisitation is prohibited. 
\item \textbf{Diff Paths Intersection by Vertex}: If enabled, distinct paths may share vertices; otherwise, vertex sharing is disallowed.
\item \textbf{Diff Paths Intersection by Edge}: If enabled, distinct paths may share edges; otherwise, edge sharing is forbidden.
\item \textbf{Final sorting mode}: Determines the method for filtering the final set of paths. The search yields two path subsets: (1) ended\_paths -- paths terminated before reaching the depth limit and (2) continuous\_paths -- paths that reached the depth limit. Each path is assigned a relevance score. The selection process varies based on the chosen mode. If ended\_first is specified, then ended\_paths are sorted in descending order by relevance and the first max\_paths paths are selected. If there are fewer ended\_paths than the max\_paths value, then continuous\_paths are sorted by relevance and the first $N$ missing paths are selected from them. If continuous\_first is specified, then paths are selected in the same way as for ended\_first, but with continuous\_paths prioritized before ended\_paths. If mixed is specified, then ended\_paths and continuous\_paths are combined into one list, sorted in descending order by relevance, and the first max\_paths paths are selected from the resulting list.
\end{enumerate}

\textbf{Mixed Algorithm.} This algorithm integrates A*, WaterCircles, and BeamSearch strategies to enhance extraction efficacy. By combining these approaches, we ensure that triples not captured by one method (e.g., WaterCircles) may still be retrieved by another (e.g., BeamSearch), thereby improving overall recall. The final set of triples, which is subsequently passed to the LLM for answer generation, constitutes the union of the outputs derived from the A*, WaterCircles, and BeamSearch algorithms.

By evaluating these diverse algorithms, this study highlights advancements in extracting pertinent information from knowledge graphs, thereby supporting the robust architectural framework required for personalizing responses in large language model (LLM) agents.
\section{Experiment Set-Up}

\subsection{Datasets}

To evaluate the proposed retrieval algorithms, we conducted experiments across three distinct benchmarks: DiaASQ \cite{li2023diaasqbenchmarkconversational}, HotpotQA, and TriviaQA. This selection was designed to evaluate our framework across varying domains, structural complexities, and reasoning requirements.

The primary evaluation dataset is DiaASQ, which consists of user dialogues from a Chinese forum focused on mobile device characteristics. A key feature of this dataset is the inclusion of structured "true statements" that encapsulate the core semantic content of each dialogue. We procedurally generated evaluation questions from these statements to ensure precise assessment. To further evaluate the framework's capability to handle temporal dynamics and contradictory information, both of which are critical for personalized agents, we extended DiaASQ with explicit temporal annotations and internally contradictory statements.

To ensure broad applicability and mitigate potential dataset bias and limited domain diversity, we supplemented DiaASQ with two widely-used, general-domain QA benchmarks:
\begin{enumerate}
    \item HotpotQA: Selected for its requirement of multi-hop reasoning across multiple documents.
    \item TriviaQA: Chosen for its factoid-style, open-domain questions that test broad knowledge retrieval.
\end{enumerate}

These datasets provide complementary challenges, moving beyond a single domain (mobile devices) to perform evaluation on general world knowledge and complex reasoning.

Considering the computational and engineering complexity of constructing and traversing large memory graphs, we created manageable yet representative subsets from the original datasets. This step was necessary to enable the iterative experimentation required for tuning multiple retrieval algorithms and LLM configurations within practical resource constraints. The preprocessing involved filtering contexts by length and, for TriviaQA, segmenting long documents into coherent chunks. The resulting subsets used for knowledge graph construction and QA pipeline evaluation are summarized in Table \ref{tab:datasets_summ}. Detailed preprocessing steps and dataset statistics are provided in Appendix~\ref{app:datasets}.

\begin{table}[ht!]
\centering
\renewcommand{\arraystretch}{1.2}
\resizebox{.35\textwidth}{!}{%
\begin{tabular}{|c|c|c|}
\hline
\textbf{Dataset} & \textbf{\#qa-pairs} & \textbf{\#contexts} \\ \hline
\textbf{DiaASQ}           & 4800                & 3483                \\ \hline
\textbf{HotpotQA}         & 2000                & 3933                \\ \hline
\textbf{TriviaQA}         & 500                 & 4925                \\ \hline
\end{tabular}%
}

\caption{Characteristics of prepared datasets for QA pipeline evaluation}
\label{tab:datasets_summ}
\end{table}

Using these benchmarks, the memorization and question-answering functionality of our PersonalAI framework is systematically compared against existing Retrieval-Augmented Generation (RAG) and GraphRAG baselines, as detailed in the subsequent sections.

\subsection{Models}

For memory graph construction (Memorize pipeline) and information retrieval (QA pipeline), we evaluate a series of 7B/8B and large-scale ($\gg$ 14B) language models to assess their performance on these tasks. The selected models include Qwen2.5 7B, DeepSeek R1 7B, Llama3.1 8B, GPT-4o-mini, and DeepSeek V3. To generate vector representations (embeddings) of natural language text data, we employ the multilingual E5-small model\footnote{https://huggingface.co/intfloat/multilingual-e5-small}.

\subsection{Graph-traversal Algorithms}

Graph-traversal algorithm evaluation is conducted for A*, WaterCircles (WC), and BeamSearch (BS), as well as their combinations: "WC + BS", "A* + BS", and "A* + WC". The values of the hyperparameters for the base algorithms were fixed (see Appendix~\ref{app:retriev_hyper}).

\subsection{Graph-traversal restrictions}

Additionally, we systematically varied the values of hyperparameters governing the traversal constraints applied to the graph during algorithm execution. These constraints determine which vertex types are excluded from traversal: "E" prohibits traversal of episodic vertices, "T" prohibits traversal of thesis vertices, and "O" prohibits traversal of object vertices. The keyword "all" means that no restrictions on graph vertex traversal are applied.

\subsection{Summary of experiment configurations}

Each QA configuration was evaluated on 100 question-answer pairs from the DiaASQ, HotpotQA, and TriviaQA datasets. The same LLM was used both for generating responses within a given QA configuration and for constructing the corresponding memory graph used to execute the QA pipeline. Consequently, for each fixed dataset/model pairing, 22 distinct QA configurations were derived. In total, 308 QA configurations were evaluated (the QA pipeline was not executed on the TriviaQA/GPT-4o-mini configurations due to resource constraints).

\subsection{Memory construction setting}

Our memory implementation consists of two main parts: a graph part and a vector part. The graph part stores textual representations of object, thesis, and episodic vertices, together with their properties and relationships (edges). Neo4j is used as the graph database for this part of the system. The vector part of memory stores vector representations (embeddings) of elements from the graph part to measure the semantic similarity of texts during QA pipeline execution. Milvus is used as the vector database for this part of the system. Our memory model also implements a caching mechanism for storing intermediate results of QA pipeline components to reduce the overall time required to process incoming questions. This component utilizes two non-relational databases: Redis and MongoDB. During our experiments, the cache was enabled. All databases were hosted and run on a single machine in personal Docker containers. Medium-sized LLMs (7B/8B) were served from a locally hosted Ollama Docker container. LLM inference during memory construction and QA pipeline processing was performed on a single NVIDIA TITAN RTX 24GB GPU.

To evaluate the QA pipeline, we constructed 14 memory graphs based on the datasets and LLMs described above. It is important to note that we disabled the stage in which outdated knowledge is searched for and deleted from memory because that functionality was outside the scope of these experiments. The average speed of adding text fragments to memory for a given database configuration is approximately 1.35 fragments per minute, with an average processed-text length of 550--650 characters. Detailed characteristics of the constructed memory graphs can be found in Appendix~\ref{app:constr_graphs}.
 
\section{Evaluation}

Traditional statistical evaluation metrics such as \textit{BLEU} \cite{papineni2002bleu}, \textit{ROUGE} \cite{lin2004rouge}, and \textit{Meteor Universal} \cite{denkowski2014meteor} struggle to distinguish syntactically similar but semantically distinct texts. While semantic methods like \textit{BERTScore} \cite{zhang2019bertscore} were introduced to address these limitations, our experiments reveal that BERTScore lacks sufficient differentiability, often failing to capture nuanced distinctions between correct and incorrect answers. Therefore, we adopt the \textit{LLM as a judge} \cite{zheng2023judging} framework and choose \textbf{Qwen2.5 7B}. The judge evaluates QA pairs using a structured prompt containing the question, ground truth and model answer. It labels $1$ for correct answers and $0$ for incorrect ones, and we use accuracy as our main metric. Corresponding LLM-prompts and details are provided in Appendix~\ref{app:judgescore_hyperp}.

Additionally, for comparison of the proposed QA pipeline with existing RAG and GraphRAG methods, the \textit{Exact Match} metric is calculated with the "ignore\_case" and "ignore\_punctuation" hyperparameters set to True.
\section{Experiments and Results}

Based on experimental results, a comparative table, summarizing best-performing QA configurations by LLM-as-a-Judge metric, was compiled (see Table~\ref{tab:qabest_retriever}). 

\begin{table*}[th!]
	\centering
	\renewcommand{\arraystretch}{1.5}
	\resizebox{\textwidth}{!}{%
		\begin{tabular}{|c|c|c|c|c|c|}
			\hline
            \begin{tabular}[c]{@{}c@{}} \textbf{LLM} / \\ \textbf{Dataset}\end{tabular} & \textbf{Qwen2.5 7B} & \textbf{DeepSeek R1 7B} & \textbf{Llama3.1 8B} & \textbf{GPT-4o-mini} & \textbf{DeepSeek V3} \\ \hline
			\textbf{DiaASQ} & \begin{tabular}[c]{@{}c@{}}0.22 / BS / all\end{tabular} & \begin{tabular}[c]{@{}c@{}}0.12 / AS / E\end{tabular} & \begin{tabular}[c]{@{}c@{}}0.19 / BS / E\end{tabular} & \begin{tabular}[c]{@{}c@{}}\underline{0.5} / BS + WC / E\end{tabular} & \begin{tabular}[c]{@{}c@{}}0.47 / BS + WC / O\end{tabular} \\ \hline
			\textbf{HotpotQA} & \begin{tabular}[c]{@{}c@{}}0.24 /  BS / O\end{tabular} & \begin{tabular}[c]{@{}c@{}}0.19 / BS / O\end{tabular} & \begin{tabular}[c]{@{}c@{}}0.47 /  BS / O\end{tabular} & \begin{tabular}[c]{@{}c@{}} \underline{0.77} / BS + WC / all\end{tabular} & \begin{tabular}[c]{@{}c@{}}0.76 /  BS + WC / T\end{tabular} \\ \hline
			\textbf{TriviaQA} & \begin{tabular}[c]{@{}c@{}}0.34 /\ BS / E\end{tabular} & \begin{tabular}[c]{@{}c@{}}0.27 / AS / E\end{tabular} & \begin{tabular}[c]{@{}c@{}}0.66 /  BS + AS / E\end{tabular} & -- & \begin{tabular}[c]{@{}c@{}}\underline{0.87} / BS + WC / all\end{tabular} \\ \hline
			Mean: & 0.27 & 0.19 & 0.44 & 0.77 & 0.70 \\ \hline
		\end{tabular}
    }
	\renewcommand{\arraystretch}{1}
\caption{\textbf{Best QA configurations ranked by the LLM-as-a-Judge metric across all experiments.} The corresponding cells contain the LLM-as-a-Judge score, the retrieval algorithm used, and the type of restriction applied to the graph during traversal. Shortcuts for retrieval algorithms: \textbf{BS} -- BeamSearch; \textbf{AS} -- A*; \textbf{BS + AS} -- hybrid of BeamSearch and A*; \textbf{BS + WC} -- hybrid of BeamSearch and WaterCircles. Shortcuts for graph restrictions: \textbf{all} -- no restrictions applied; \textbf{E} -- episodic vertices excluded from traversal; \textbf{T} -- thesis vertices excluded; \textbf{O} -- object vertices excluded.}
	\label{tab:qabest_retriever}
\end{table*}

As shown in Table \ref{tab:qabest_retriever}, Qwen2.5 achieved the best performance (0.27) among 7B models. Among all evaluated configurations, the highest overall effectiveness (0.77) was reached by setups incorporating GPT-4o-mini. Notably, the top-performing 7B configurations predominantly relied on BeamSearch, especially under the constraint that traversal through episodic vertices was restricted. In contrast, the best DeepSeek V3 configurations frequently adopted a hybrid strategy combining BeamSearch and WaterCircles. Across high-performing configurations more broadly, BeamSearch consistently appeared as a key component of the retrieval pipeline.

To evaluate the effect of imposed constraints on the quality of the QA pipeline, we construct two distinct distributions of values, each corresponding to a specific model/dataset pair and retrieval algorithm. These distributions were then averaged across datasets and models for configurations exhibiting the lowest and highest LLM-as-a-Judge values, as detailed in Table \ref{tab:qaweak_nodestypes}.

\begin{table*}[!ht]
	\centering
	\renewcommand{\arraystretch}{1.5}
	\resizebox{\textwidth}{!}{%
		\begin{tabular}{|c|cc|cc|cc|}
			\hline
			\multirow{2}{*}{\begin{tabular}[c]{@{}c@{}}\textbf{Vertex types} \\ \textbf{restrictions}\end{tabular}} & \multicolumn{2}{c|}{\textbf{7B}} & \multicolumn{2}{c|}{\textbf{8B}} & \multicolumn{2}{c|}{\textbf{14B+}} \\ \cline{2-7} 
			& \multicolumn{1}{c|}{\textbf{worse configs}} & \textbf{best configs} & \multicolumn{1}{c|}{\textbf{worse configs}} & \textbf{best configs} & \multicolumn{1}{c|}{\textbf{worse configs}} & \textbf{best configs} \\ \hline
			\textbf{E} & \multicolumn{1}{c|}{3\%} & \underline{44\%} & \multicolumn{1}{c|}{9\%} & \underline{45\%} & \multicolumn{1}{c|}{27\%} & 7\% \\ \hline
			\textbf{T} & \multicolumn{1}{c|}{\underline{84\%}} & 12\% & \multicolumn{1}{c|}{\underline{64\%}} & 31\% & \multicolumn{1}{c|}{20\%} & \underline{73\%} \\ \hline
			\textbf{O} & \multicolumn{1}{c|}{13\%} & \underline{44\%} & \multicolumn{1}{c|}{27\%} & 25\% & \multicolumn{1}{c|}{\underline{53\%}} & 20\% \\ \hline
		\end{tabular}%
	}
	\renewcommand{\arraystretch}{1}
\caption{Impact of various constraints imposed during memory graph traversal on the quality of the QA pipeline: "worse configs" -- distribution for low-quality configurations; "best configs" -- distribution for high-quality configurations}
	\label{tab:qaweak_nodestypes}
\end{table*}

As demonstrated in Table \ref{tab:qaweak_nodestypes}, for 7B/8B models, the majority ($\approx$74\%) of configurations yielding the lowest response quality impose restrictions on traversing thesis-type vertices. Conversely, a significant proportion of high-quality configurations restrict traversal of episodic and object vertices ($\approx$44\% and $\approx$34\%, respectively). This suggests that thesis-type memories contain critical information for generating relevant responses, whereas the inclusion of episodic and object memories introduces noise into the context, thereby degrading output quality. For larger-scale models, the trend differs: 53\% of low-quality configurations restrict traversal of object vertices, while 73\% of high-quality configurations restrict traversal of thesis-type vertices. This implies that larger models exhibit greater robustness in handling conditional generation from lengthy or noisy episodic memories, rendering thesis-based information redundant.

\begin{table*}[!ht]
	\centering
	\renewcommand{\arraystretch}{1.5}
	\resizebox{0.92\textwidth}{!}{%
		\makebox[\textwidth]{%
			\rule{0cm}{1mm}
		\begin{tabular}{|c|ccc|ccc|ccc|}
			\hline
			\multirow{3}{*}{\begin{tabular}[c]{@{}c@{}}\textbf{Retrieval} \\ \textbf{algorithm}\end{tabular}} & \multicolumn{3}{c|}{\textbf{7B}} & \multicolumn{3}{c|}{\textbf{8B}} & \multicolumn{3}{c|}{\textbf{14B+}} \\ \cline{2-10} 
			& \multicolumn{1}{c|}{\begin{tabular}[c]{@{}c@{}}\textbf{worse configs} \\ \textbf{(w restr)}\end{tabular}} & \multicolumn{1}{c|}{\begin{tabular}[c]{@{}c@{}}\textbf{best configs} \\ \textbf{(w restr)}\end{tabular}} & \begin{tabular}[c]{@{}c@{}}\textbf{other configs} \\ \textbf{(w/o restr)}\end{tabular} & \multicolumn{1}{c|}{\begin{tabular}[c]{@{}c@{}}\textbf{worse configs} \\ \textbf{(w restr)}\end{tabular}} & \multicolumn{1}{c|}{\begin{tabular}[c]{@{}c@{}}\textbf{best configs} \\ \textbf{(w restr)}\end{tabular}} & \begin{tabular}[c]{@{}c@{}}\textbf{other configs} \\ \textbf{(w/o restr)}\end{tabular} & \multicolumn{1}{c|}{\begin{tabular}[c]{@{}c@{}}\textbf{worse configs} \\ \textbf{(w restr)}\end{tabular}} & \multicolumn{1}{c|}{\begin{tabular}[c]{@{}c@{}}\textbf{best configs} \\ \textbf{(w restr)}\end{tabular}} & \begin{tabular}[c]{@{}c@{}}\textbf{other configs} \\ \textbf{(w/o restr)}\end{tabular} \\ \hline
			\textbf{WC} & \multicolumn{1}{c|}{--} & \multicolumn{1}{c|}{--} & 0.09 & \multicolumn{1}{c|}{--} & \multicolumn{1}{c|}{--} & 0.34 & \multicolumn{1}{c|}{--} & \multicolumn{1}{c|}{--} & 0.55 \\ \hline
			\textbf{AS} & \multicolumn{1}{c|}{\underline{0.1}} & \multicolumn{1}{c|}{0.175} & \underline{0.14} & \multicolumn{1}{c|}{0.29} & \multicolumn{1}{c|}{0.36} & 0.41 & \multicolumn{1}{c|}{0.23} & \multicolumn{1}{c|}{0.36} & 0.33 \\ \hline
			\textbf{BS} & \multicolumn{1}{c|}{0.025} & \multicolumn{1}{c|}{\underline{0.18}} & 0.06 & \multicolumn{1}{c|}{0.26} & \multicolumn{1}{c|}{\underline{0.5}} & 0.36 & \multicolumn{1}{c|}{0.48} & \multicolumn{1}{c|}{0.6} & 0.65 \\ \hline
			\textbf{WC+BS} & \multicolumn{1}{c|}{0.033} & \multicolumn{1}{c|}{0.095} & 0.02 & \multicolumn{1}{c|}{0.30} & \multicolumn{1}{c|}{0.39} & 0.32 & \multicolumn{1}{c|}{\underline{0.62}} & \multicolumn{1}{c|}{\underline{0.7}} & \underline{0.68} \\ \hline
			\textbf{BS+AS} & \multicolumn{1}{c|}{0.02} & \multicolumn{1}{c|}{0.175} & 0.01 & \multicolumn{1}{c|}{0.25} & \multicolumn{1}{c|}{0.48} & 0.36 & \multicolumn{1}{c|}{0.48} & \multicolumn{1}{c|}{0.64} & 0.66 \\ \hline
			\textbf{AS+WC} & \multicolumn{1}{c|}{0.055} & \multicolumn{1}{c|}{0.115} & 0.07 & \multicolumn{1}{c|}{\underline{0.33}} & \multicolumn{1}{c|}{0.37} & \underline{0.42} & \multicolumn{1}{c|}{0.57} & \multicolumn{1}{c|}{0.6} & 0.6 \\ \hline
            Mean: & \multicolumn{1}{c|}{0.046} & \multicolumn{1}{c|}{0.148} & 0.065 & \multicolumn{1}{c|}{0.286} & \multicolumn{1}{c|}{0.42} & 0.36 & \multicolumn{1}{c|}{0.47} & \multicolumn{1}{c|}{0.58} & 0.57 \\ \hline
		\end{tabular}%
	}}
	\renewcommand{\arraystretch}{1}
	\caption{Stability of proposed retrieval algorithms when various restrictions are imposed: "worse--configs (w restr)" -- low LLM-as-a-Judge score configurations and imposed restrictions; "best--configs (w restr)" -- high LLM-as-a-Judge score configurations and imposed restrictions; "other--configs (w/o restr)" -- configurations without restrictions on graph traversal}
	\label{tab:qaweak_bestsliceretr}
\end{table*}

Additionally, the comparative analysis presented in Table~\ref{tab:qaweak_bestsliceretr} evaluates the robustness of various graph traversal algorithms. The results indicate that configurations employing 8B models in conjunction with a combined search strategy (A* and WaterCircles) demonstrate high stability, with performance degradation remaining within 4\% across varying traversal constraints. In contrast, the BeamSearch algorithm exhibits high sensitivity to these constraints: suboptimal parameterization results in substantial performance loss, with the LLM-as-a-Judge score varying by as much as 24\% between optimal and non-optimal settings. However, for larger-scale models, the combination of BeamSearch and WaterCircles yields more consistent performance, suggesting improved robustness at higher model capacities.

A critical component of the implemented QA pipeline is the "NoAnswer" mechanism. This mechanism incorporates a directive in the LLM prompt, instructing the model to output a predefined symbol if provided context lacks sufficient information to generate a valid response. Table \ref{tab:qaweak_noanswer} summarizes the frequency of "NoAnswer" outputs across different model configurations, retrieval algorithms, and graph traversal constraints.

\begin{table*}[!ht]
\centering
\renewcommand{\arraystretch}{1.5}
\resizebox{\textwidth}{!}{%
\begin{tabular}{|c|c|c|c|c|c|c|cc|c|c|c|c|}
\cline{1-7} \cline{9-13}
\begin{tabular}[c]{@{}c@{}}\textbf{Retrieval} \\ \textbf{algorithm /}\\ \textbf{LLM size}\end{tabular} & \textbf{WC} & \textbf{AS} & \textbf{BS} & \textbf{WC+BS} & \textbf{BS+AS} & \textbf{AS+WC} & \multicolumn{1}{c|}{} & \begin{tabular}[c]{@{}c@{}}\textbf{Vertex types}\\ \textbf{restrictions /}\\ \textbf{LLM size}\end{tabular} & \textbf{all} & \textbf{E} & \textbf{T} & \textbf{O} \\ \cline{1-7} \cline{9-13} 
\textbf{7B} & 31\% & 44\% & 26\% & 29\% & 31\% & \underline{25\%} &  &  & 33\% & 27\% & \underline{26\%} & 36\% \\ \cline{1-7} \cline{10-13} 
\textbf{8B} & 51\% & 49\% & \underline{43\%} & 51\% & 73\% & 49\% &  &  & 51\% & \underline{40\%} & 51\% & 46\% \\ \cline{1-7} \cline{10-13} 
\textbf{14B+} & 25\% & 62\% & 27\% & \underline{16\%} & 26\% & 21\% &  &  & \underline{26\%} & 32\% & 27\% & 35\% \\ \cline{1-7} \cline{10-13} 
\end{tabular}%
}
\renewcommand{\arraystretch}{1}
\caption{The influence of selected retrieval algorithms and imposed search restrictions on the percentage of generated "NoAnswer" stubs}
\label{tab:qaweak_noanswer}
\end{table*}

As demonstrated in Table \ref{tab:qaweak_noanswer}, the analysis reveals distinct patterns in the occurrence of "NoAnswer" responses across different model configurations. For 7B models, the lowest frequency of "NoAnswer" responses is observed when employing a combined A* and WaterCircles algorithm with restricted traversal of thesis vertices. In contrast, 8B models exhibit minimal "NoAnswer" instances when utilizing the BeamSearch algorithm alongside a prohibition on episodic-vertex traversal. For larger-scale models, the optimal performance, measured by the fewest "NoAnswer" responses, is achieved through a combined BeamSearch and WaterCircles approach without graph traversal constraints. These findings suggest that the aforementioned algorithms are more effective at extracting relevant information than alternative methods under the specified conditions.

It is also important to note the time required to process a single user question with the QA pipeline for a given LLM and retrieval algorithm; see Table \ref{tab:qapipe_latency}.

\begin{table*}[ht!]
\centering
\renewcommand{\arraystretch}{1.5}
\resizebox{\textwidth}{!}{%
\begin{tabular}{|c|ccccc|l|}
\hline
\multirow{3}{*}{\begin{tabular}[c]{@{}c@{}}\textbf{LLM /} \\ \textbf{Retrieval} \\ \textbf{algorithm}\end{tabular}} & \multicolumn{5}{c|}{\textbf{QA pipeline Latency (minutes)}} & \multirow{3}{*}{Mean:} \\ \cline{2-6}
 & \multicolumn{1}{c|}{\multirow{2}{*}{\textbf{Qwen2.5 7B}}} & \multicolumn{1}{c|}{\multirow{2}{*}{\textbf{DeepSeek R1 7B}}} & \multicolumn{1}{c|}{\multirow{2}{*}{\textbf{Llama3.1 8B}}} & \multicolumn{1}{c|}{\multirow{2}{*}{\textbf{GPT4o mini}}} & \multirow{2}{*}{\textbf{DeepSeek V3}} &  \\
 & \multicolumn{1}{c|}{} & \multicolumn{1}{c|}{} & \multicolumn{1}{c|}{} & \multicolumn{1}{c|}{} &  &  \\ \hline
\textbf{WC} & \multicolumn{1}{c|}{0.14} & \multicolumn{1}{c|}{0.34} & \multicolumn{1}{c|}{0.46} & \multicolumn{1}{c|}{0.22} & 0.33 & \multicolumn{1}{c|}{0.30} \\ \hline
\textbf{AS} & \multicolumn{1}{c|}{2.24} & \multicolumn{1}{c|}{4.68} & \multicolumn{1}{c|}{3.51} & \multicolumn{1}{c|}{-} & 2.53 & \multicolumn{1}{c|}{3.24} \\ \hline
\textbf{BS} & \multicolumn{1}{c|}{5.08} & \multicolumn{1}{c|}{7.86} & \multicolumn{1}{c|}{5.00} & \multicolumn{1}{c|}{8.70} & 6.32 & \multicolumn{1}{c|}{6.59} \\ \hline
\end{tabular}%
}
\caption{QA pipeline latency (in minutes) as a function of the LLM and retrieval algorithm, with the caching mechanism enabled}
\label{tab:qapipe_latency}
\end{table*}

Table \ref{tab:qapipe_latency} shows that QA pipelines using the WaterCircles retriever are significantly faster. This is because WaterCircles does not need to use the vector component of the memory graph, which stores vector representations of its elements, to perform traversal. Conversely, the QA pipeline with the BeamSearch algorithm turned out to be slower than the QA pipeline with the A* algorithm. This is because A* requires embeddings of memory graph elements to construct and traverse only one path on the graph, while BeamSearch must construct and monitor $N$ candidate paths to select the optimal traversal. The observed latency of the traversal algorithms is also highly dependent on the chosen vector database when configuring the vector component of memory. In our experiments, Milvus was used. After completing our main experiments, we evaluated the read/write performance of five databases (Milvus, OpenSearch, Weaviate, Elasticsearch, and Qdrant) and found that (1) Qdrant was the fastest database, and (2) Qdrant was six times faster than Milvus. Therefore, we recommend using Qdrant when configuring our memory graph to reduce the average time required to process a single question with the proposed QA pipeline.

In summary, our framework demonstrates that accuracy can be improved by configuring the memory graph ontology and retrieval methods according to the available LLM and the selected QA task. A comparative analysis of existing RAG and GraphRAG methods against our proposed approach, conducted on the TriviaQA and HotpotQA datasets, is provided in Appendix~\ref{app:rag_graphrag_cmpr}.

\section{Conclusion}
This work introduces a flexible and extensible framework for integrating structured memory into language model agents based on a knowledge graph. By extending the AriGraph architecture with support for object, episodic, and thesis vertices, as well as hyper-edges, we enable rich temporal and semantic representations that go beyond traditional RAG pipelines. Our system supports multiple retrieval algorithms, including A*, WaterCircles, BeamSearch, and hybrid combinations that can be dynamically adapted to the model’s scale and task requirements.
Through extensive evaluation on three benchmarks (DiaASQ, HotpotQA, and TriviaQA), we demonstrate that performance varies systematically with the choice of retrieval strategy and graph traversal constraints. For smaller-scale models (7B–8B), configurations that restrict episodic or object vertices and rely on BeamSearch yield the highest accuracy, while for larger models, hybrid methods combining BeamSearch and WaterCircles offer improved stability and robustness. Importantly, we show that thesis vertices often encode critical information, and excluding them typically degrades performance, especially in 7B models.
Compared to existing RAG and GraphRAG methods, our approach demonstrates competitive or superior performance, particularly in handling temporally complex and contradictory information. Ablation studies further reveal that hybrid traversal strategies reduce sensitivity to graph constraints and lower the frequency of invalid responses.
Overall, our system provides a principled architecture for long-term, structured memory in LLM agents, enabling personalized, context-aware reasoning at scale. It lays the groundwork for future extensions involving temporal filtering, edge-type prioritization, and more fine-grained memory control.

\section{Future work}
In future work, we first propose to enhance the temporal dynamics of our memory graph by introducing a "memory time" parameter, which will enable fine-grained filtering of triples based on temporal proximity and edge types. This modification will allow the system to selectively prioritize temporally proximate data or emphasize specific relationship categories, thereby improving the precision of personalized responses. Second, recognizing potential bottlenecks associated with the current implementation of graph traversal algorithms, we will focus on fine-tuning the underlying vector storage schemes employed alongside advanced approximate nearest neighbor search techniques. These enhancements promise substantial reductions in overall query latency while maintaining comparable precision rates. Third, to reduce the vector search space and speed up vector retrieval operations, we plan to add more characteristics by which triples from the knowledge graph can be aggregated and stored in separate, smaller, but more concentrated and specific vector stores.

Also, one possible future direction is to explore erasure-coded and locally recoverable layouts for sharding graph and vector indices across nodes, inspired by information-theoretic distributed storage, enabling fast repair and continued operation under partial server unavailability \cite{10.1134/S1064226920120116}. In addition, we will investigate private and verifiable retrieval protocols (PIR with result verification) so an agent can query remote memory without revealing the user’s intent and can detect incorrect or malicious responses \cite{10.48550/arXiv.2301.11730}. These mechanisms aim to make long-term personalized memory robust, secure, and auditable at scale. 

\bibliographystyle{unsrt}
\bibliography{references}
\vspace*{-20mm}

\begin{IEEEbiography}[{\includegraphics[width=1in,height=1.25in,clip,keepaspectratio]{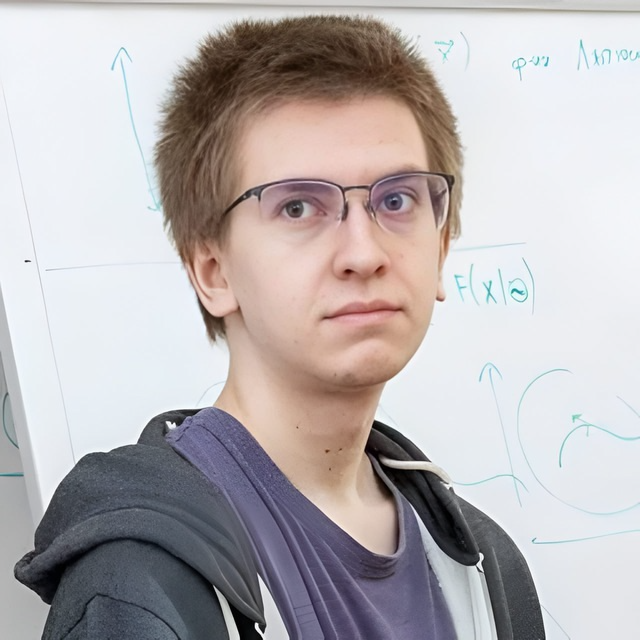}}]{M. Menschikov} received the B.Sc. degree in Software Engineering from Petrozavodsk State University in 2023 and the M.Sc. degree in Machine Learning Engineering from ITMO University in 2025. He is currently a Software Engineer at the Skoltech AI Center, where he contributed to a project on developing working memory for LLM agents based on a knowledge graph. His research interests include generative modeling, GraphRAG, LLM-based knowledge graph reasoning, LLM-based knowledge graph construction, multi-agent systems, and dialogue systems.
\end{IEEEbiography}

\vspace*{-18mm}

\begin{IEEEbiography}[{\includegraphics[width=1in,height=1.25in,clip,keepaspectratio]{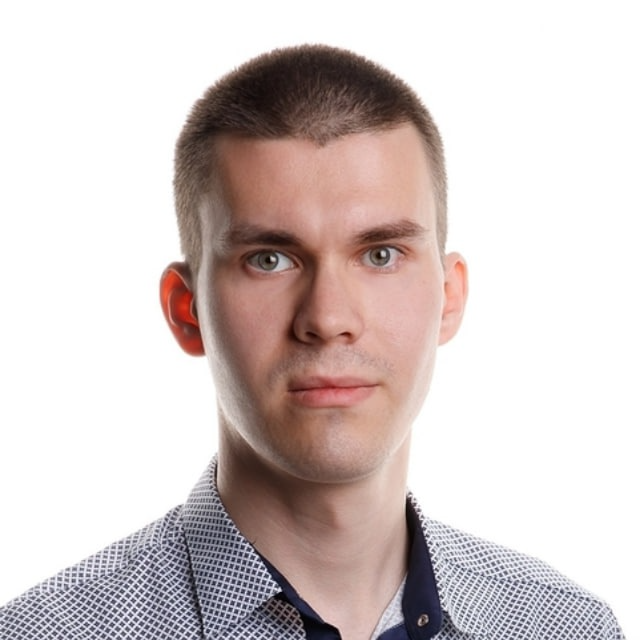}}]{D. Evseev} completed a master's degree and postgraduate study at the Moscow Institute of Physics and Technology (the Phystech School of Applied Mathematics and Computer Science), where he later received the Ph.D. degree. He is currently a Senior Research Engineer at Skoltech.
\end{IEEEbiography}

\vspace*{-13mm}

\begin{IEEEbiography}[{\includegraphics[width=1in,height=1.25in,clip,keepaspectratio]{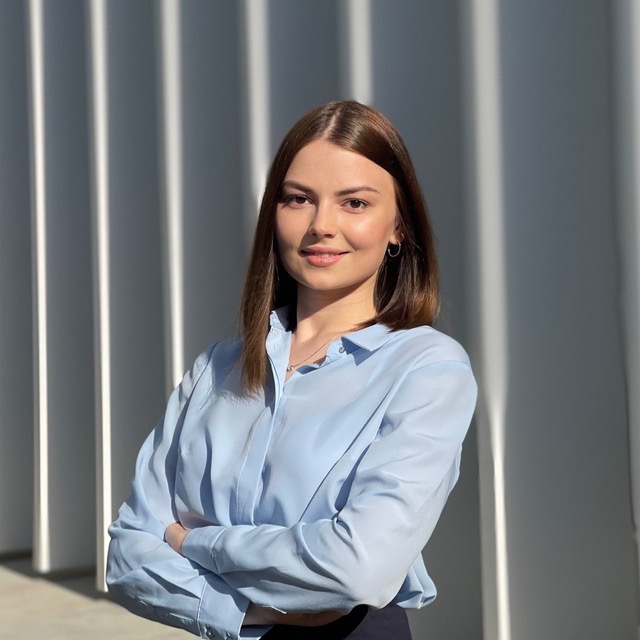}}]{V. Dochkina} is the Director of the AI and Data Center, Strategy and Development Block, Sber. She leads the block’s strategic AI implementation and digital transformation initiatives.
Education: Bachelor’s and Master’s degrees with honors from the Moscow Institute of Physics and Technology (MIPT); Master’s degree from Skoltech, recipient of the 2021 Best Thesis Award; ongoing PhD at MIPT focused on multiagent AI systems and foundation model architectures.
Expertise: Development and deployment of enterprise-scale AI solutions; AI governance frameworks; Agentic AI.
Research interests: Foundation models; multimodal expansion; agentic LLM capability development; scaling AI agents for process automation; autonomous AI systems; mixture-of-experts architectures; coordination frameworks for enterprise-wide autonomization.
\end{IEEEbiography}

\vspace*{-13mm}

\begin{IEEEbiography}[{\includegraphics[width=1in,height=1.25in,clip,keepaspectratio]{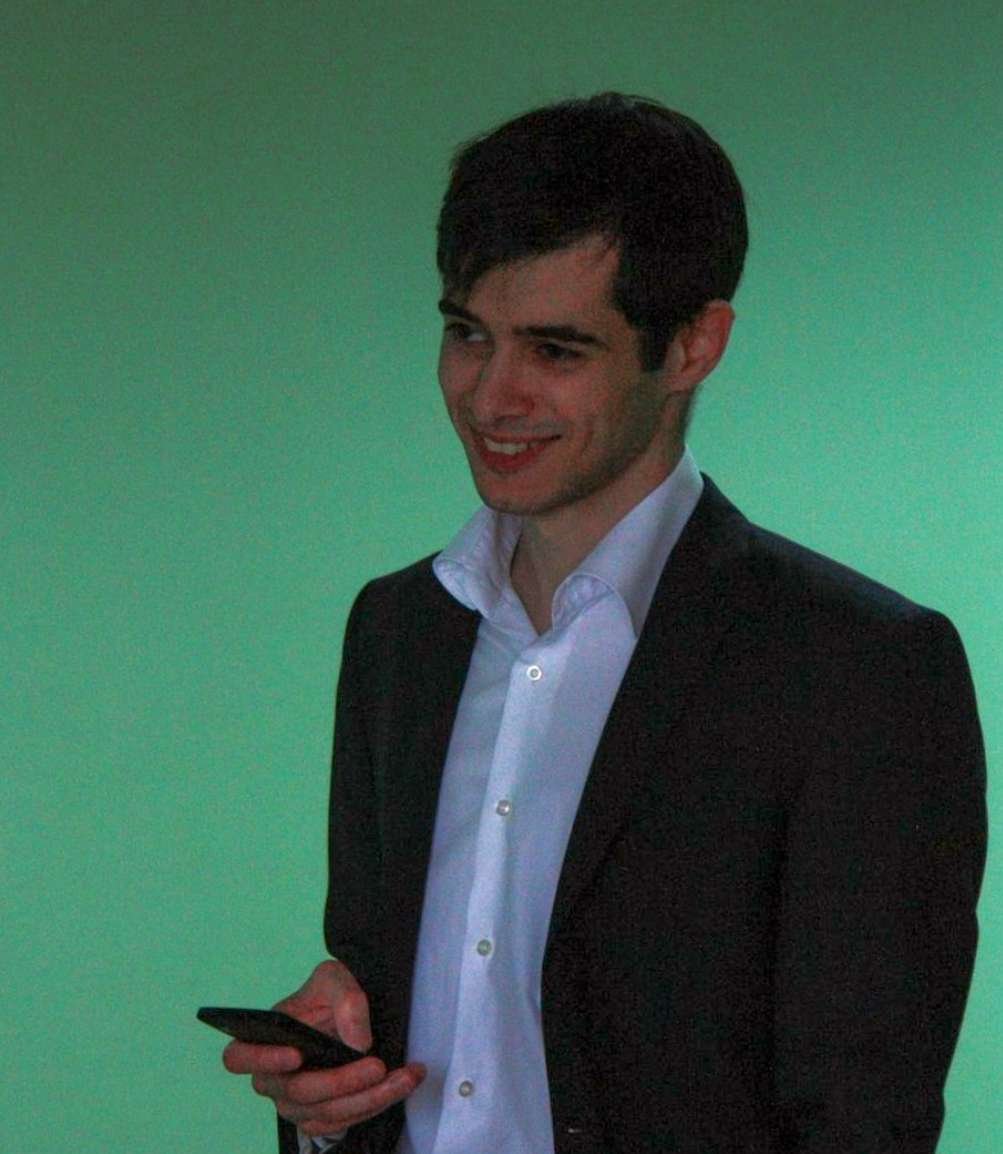}}]{R. Kostoev} got M.Sc. degree in applied mathematics and computer science from Lomonosov Moscow State University, and built an impressive career spanning technology, innovation, and leadership roles. His professional journey includes experience at major companies such as Philips and Google, where he contributed to significant projects and initiatives. 
\end{IEEEbiography}

\vspace*{-13mm}

\begin{IEEEbiography}[{\includegraphics[width=1in,height=1.25in,clip,keepaspectratio]{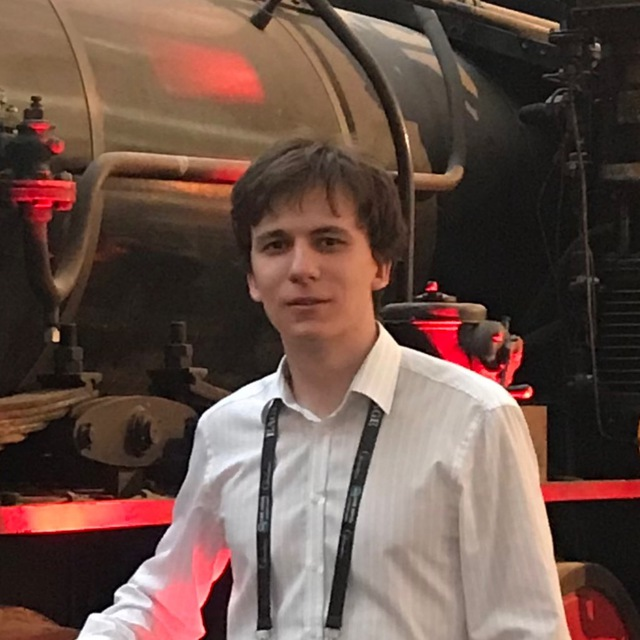}}]{I. Perepechkin} got M.Sc. degree in Applied Mathematics and Physics from Moscow Institute of Physics and Technology in 2017. He has experience developing enterprise-level AI solutions. He is currently a team lead data scientist at Sberbank, developing multi-agent systems.
\end{IEEEbiography}

\vspace*{-13mm}

\begin{IEEEbiography}[{\includegraphics[width=1in,height=1.25in,clip,keepaspectratio]{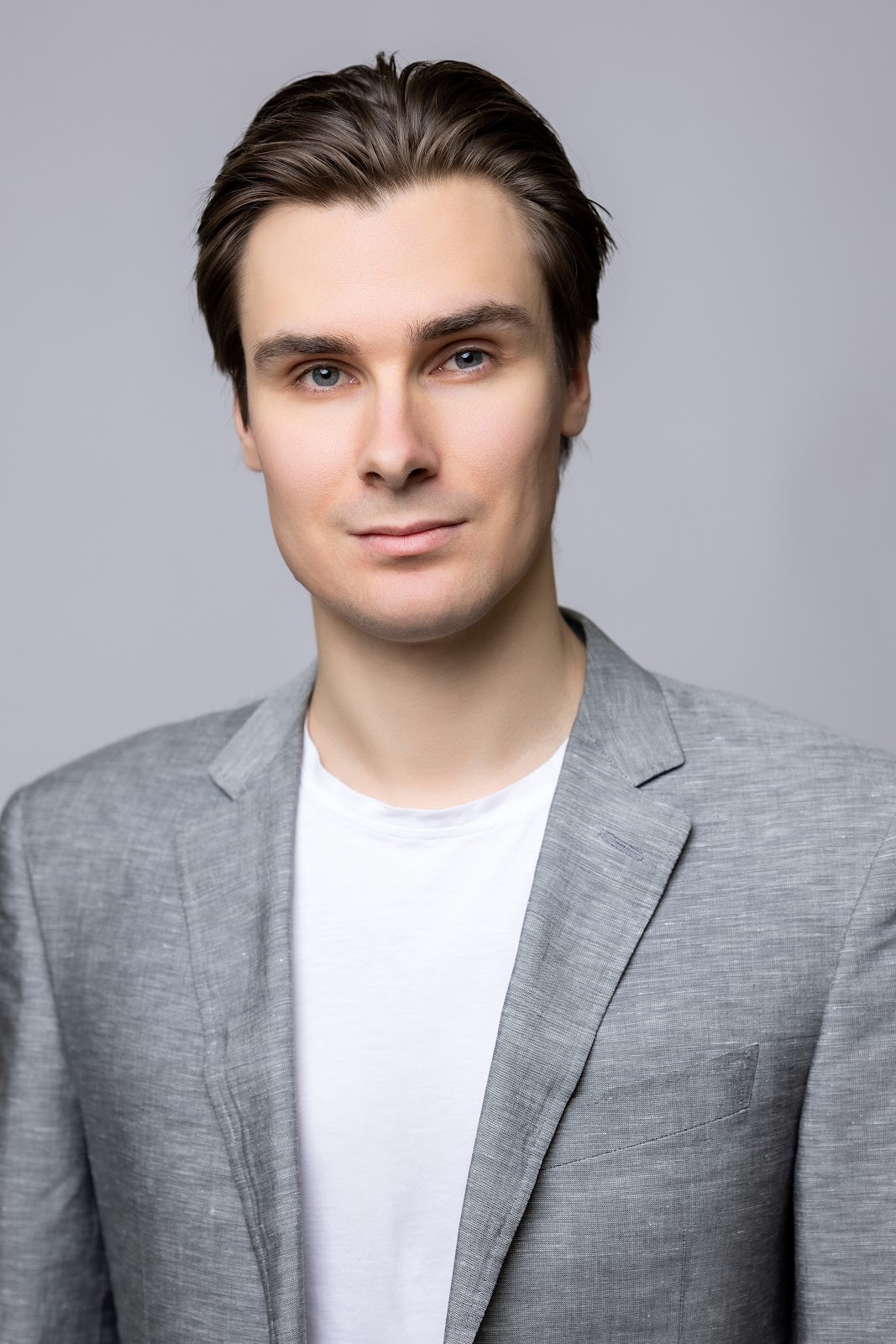}}]{P. Anokhin} received his M.Sc. degree in Physiology from Lomonosov Moscow State University in 2013 and his Ph.D. in Physiology and Biochemistry from the Russian Academy of Sciences in 2017. His early research focused on studying the dopamine system in animal models of addiction and reinforcement learning. In 2021, he joined AIRI, where he now leads a team researching large language model (LLM) agents, reasoning models, and memory architectures for intelligent agents.
\end{IEEEbiography}

\vspace*{-13mm}

\begin{IEEEbiography}[{\includegraphics[width=1in,height=1.25in,clip,keepaspectratio]{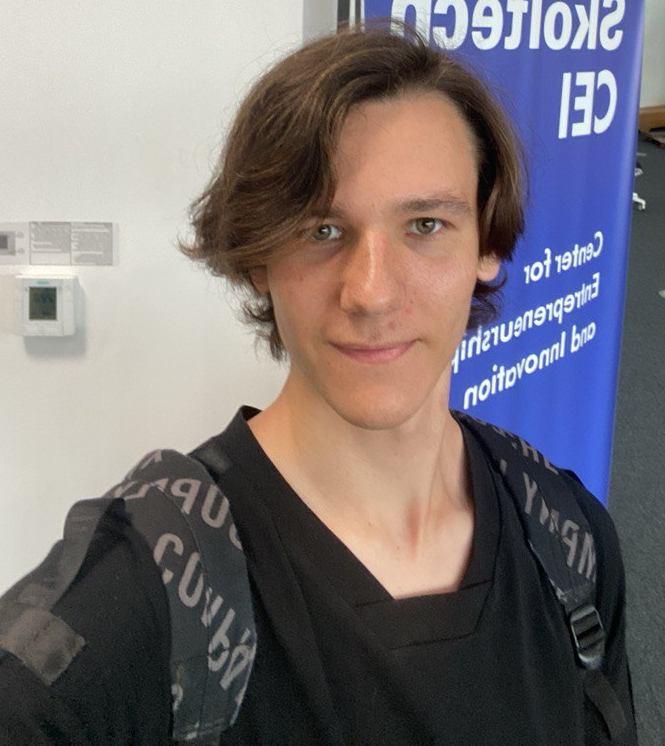}}]{N. Semenov} studied for three years at the Faculty of Mechanics and Mathematics of Lomonosov Moscow State University, then transferred to the Faculty of Mathematics at the Higher School of Economics, where he completed his bachelor’s degree. He is currently not affiliated with any research institution and is engaged in independent research.
\end{IEEEbiography}

\begin{IEEEbiography}[{\includegraphics[width=1in,height=1.25in,clip,keepaspectratio]{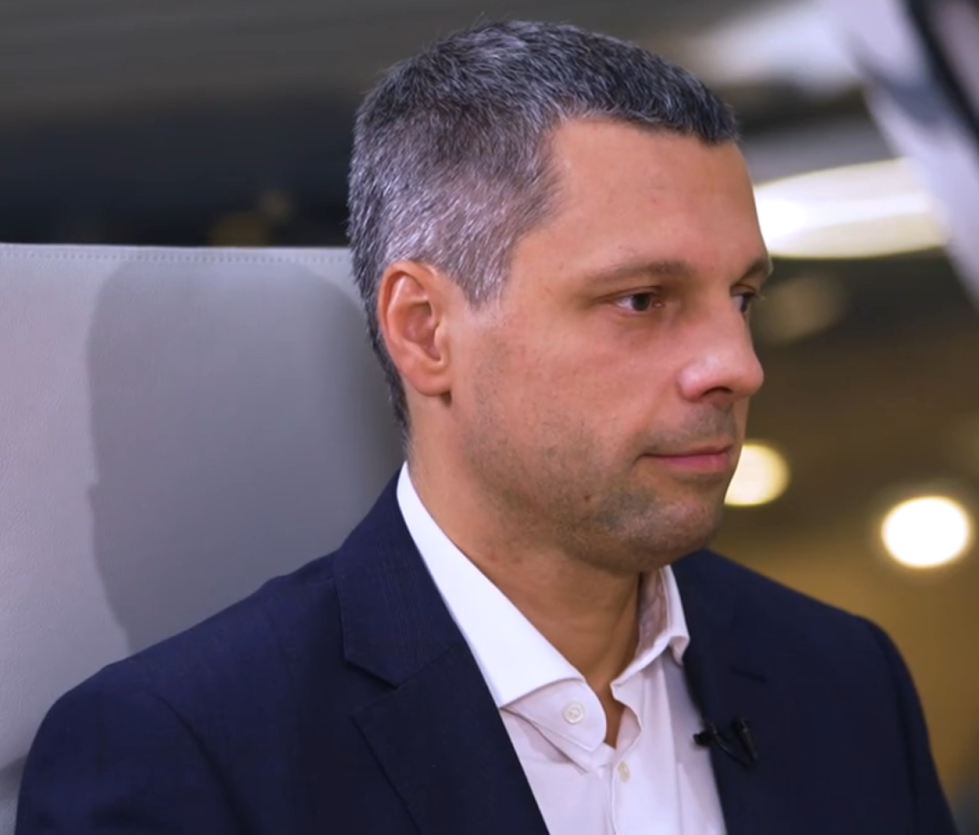}}]{E. Burnaev} received the M.Sc. degree in applied physics and mathematics from Moscow Institute of Physics and Technology, in 2006, the Ph.D. degree in foundations of computer science from the Institute for Information Transmission Problem RAS, in 2008, and the Dr.Sci. degree in mathematical modeling and numerical methods from Moscow Institute of Physics and Technology, in 2022. 
He is currently the Director of the AI Center, Skolkovo Institute of Science and Technology, and a Full Professor. His research interests include generative modeling, manifold learning, deep learning for 3D data analysis, multi-agent systems, and industrial applications.
\end{IEEEbiography}

\vspace*{5mm}

\clearpage
\onecolumn
\appendices

\section{LLM prompts used to build memory graph by Memorize pipeline}
\label{app:mem_add_prompts}

Tables \ref{tab:mempropmt_ethesis} and \ref{tab:mempropmt_esimple} present the LLM prompts employed in the Memorize pipeline for extracting thesis and object memory triples, respectively, from unstructured natural language text. These prompts facilitate the transformation of textual data into a structured knowledge graph representation.

\begin{table}[H]
	\centering
	\resizebox{\textwidth}{!}{%
		\begin{tabular}{|c|l|}
			\hline
			\textbf{Type} & \multicolumn{1}{c|}{\textbf{Prompt}} \\ \hline
			\multirow{1}{*}{\textbf{System}} & 
			\begin{minipage}{\textwidth}
				\includegraphics[clip,trim={.07\textwidth} {.6\textheight} {.08\textwidth} 12mm, width=\textwidth,valign=b]{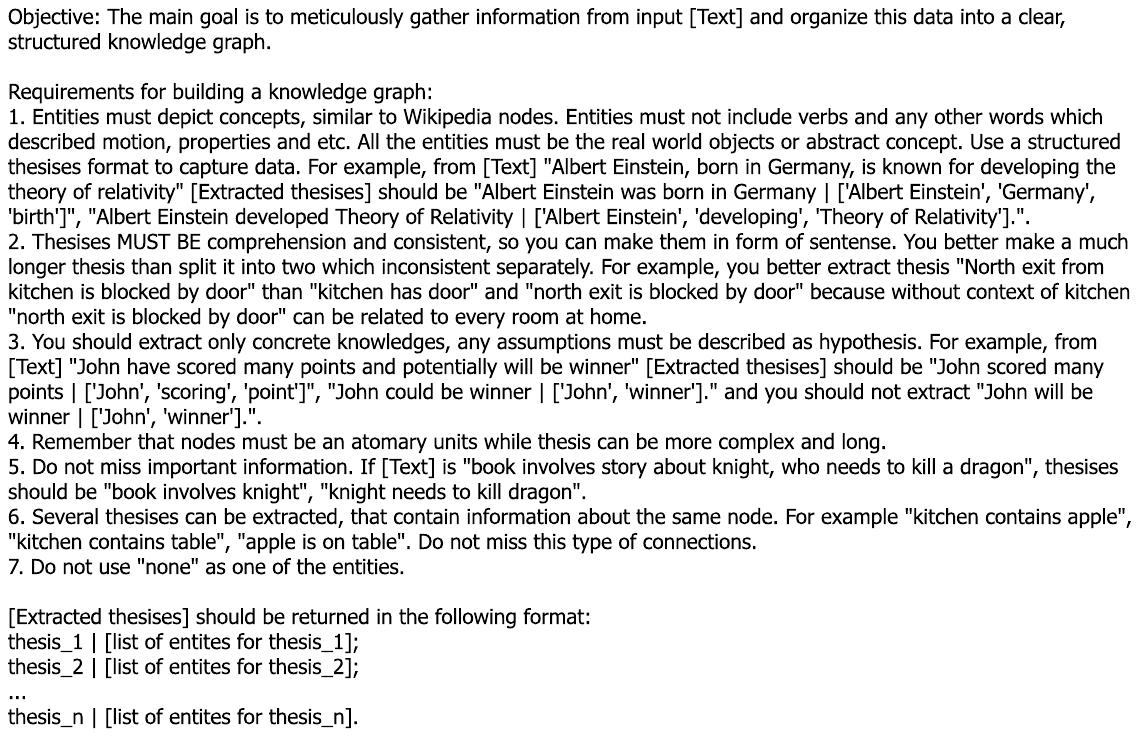}
			\end{minipage}
			\\ \hline
			
			\multirow{1}{*}{\textbf{User}} & 
			\begin{minipage}{\textwidth}
				\includegraphics[clip,trim={.07\textwidth} {\textheight} {.08\textwidth} 12mm, width=\textwidth,valign=b]{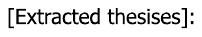}
			\end{minipage}
			\\ \hline

			\multirow{1}{*}{\textbf{Assistant}} & 
			\begin{minipage}{\textwidth}
				\includegraphics[clip,trim={.07\textwidth} {\textheight} {.08\textwidth} 12mm, width=\textwidth,valign=b]{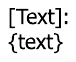}
			\end{minipage}
			\\ \hline
		\end{tabular}%
	}
	\caption{LLM prompts for extracting thesis memories (in the form of triples) from natural language text}
	\label{tab:mempropmt_ethesis}
\end{table}

\begin{table}[H]
	\centering
	\resizebox{\textwidth}{!}{%
		\begin{tabular}{|c|l|}
			\hline
			\textbf{Type} & \multicolumn{1}{c|}{\textbf{Prompt}} \\ \hline
			\multirow{1}{*}{\textbf{System}} & 
			\begin{minipage}{\textwidth}
				\includegraphics[clip,trim={.07\textwidth} {.44\textheight} {.08\textwidth} 12mm, width=\textwidth,valign=b]{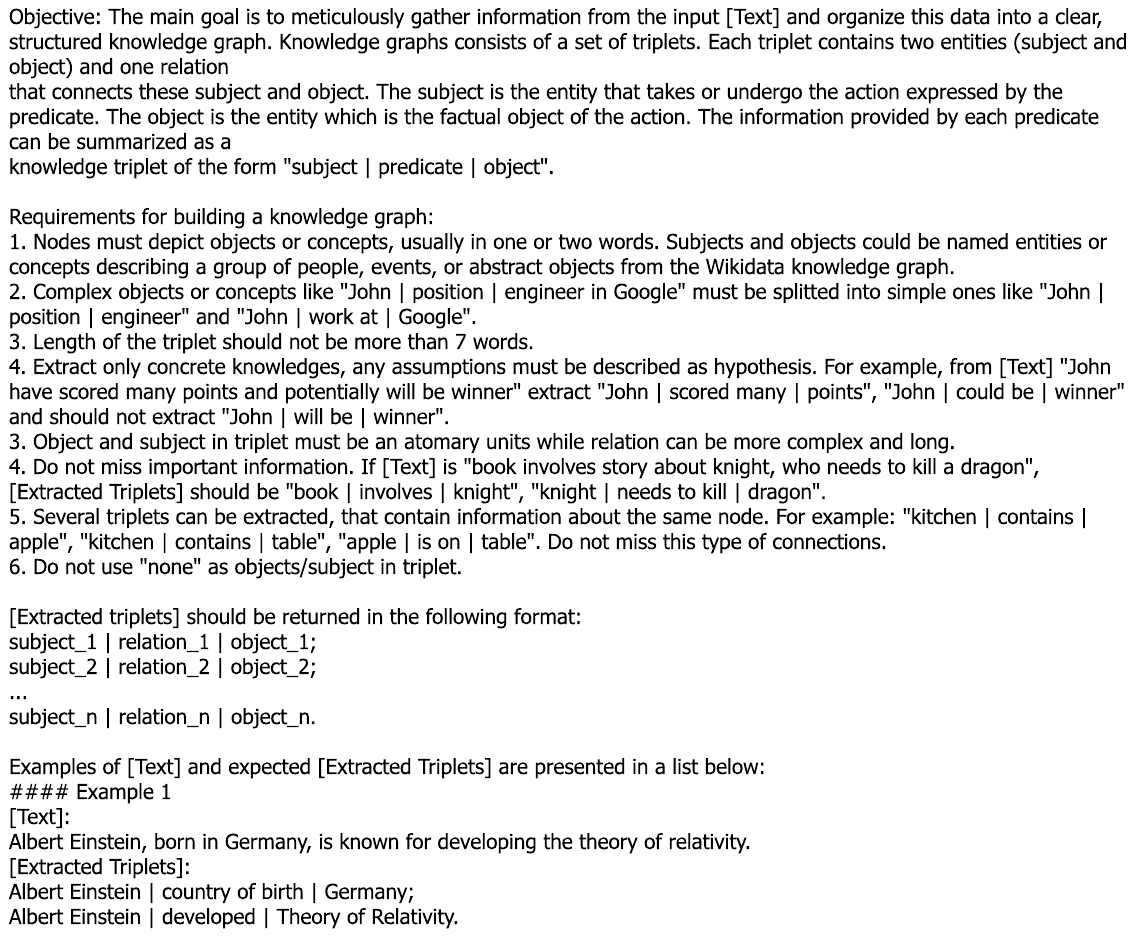}
			\end{minipage}
			\\ \hline
						
			\multirow{1}{*}{\textbf{User}} & 
			\begin{minipage}{\textwidth}
				\includegraphics[clip,trim={.07\textwidth} {\textheight} {.08\textwidth} 12mm, width=\textwidth,valign=b]{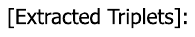}
			\end{minipage}
			\\ \hline
		
			\multirow{1}{*}{\textbf{Assistant}} & 
			\begin{minipage}{\textwidth}
				\includegraphics[clip,trim={.07\textwidth} {\textheight} {.08\textwidth} 12mm, width=\textwidth,valign=b]{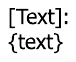}
			\end{minipage}
			\\ \hline
		\end{tabular}%
	}
	\caption{LLM prompts for extracting object memories (in the form of triples) from natural language text}
	\label{tab:mempropmt_esimple}
\end{table}

\section{LLM prompts used to find outdated information in constructed memory graph}
\label{app:mem_upd_prompts}

Tables \ref{tab:mempropmt_rthesis} and \ref{tab:mempropmt_rsimple} displays LLM prompts employed in Memorize pipeline for identifying stale thesis-related and object memories, respectively, in memory graph.

\begin{table}[H]
	\centering
	\resizebox{\textwidth}{!}{%
		\begin{tabular}{|c|l|}
			\hline
			\textbf{Type} & \multicolumn{1}{c|}{\textbf{Prompt}} \\ \hline
			\multirow{1}{*}{\textbf{System}} & 
			\begin{minipage}{\textwidth}
				\includegraphics[clip,trim={.07\textwidth} {\textheight} {.08\textwidth} 12mm, width=\textwidth,valign=b]{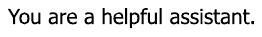}
			\end{minipage}
			\\ \hline
			
			\multirow{1}{*}{\textbf{User}} & 
			\begin{minipage}{\textwidth}
				\includegraphics[clip,trim={.07\textwidth} {.14\textheight} {.08\textwidth} 12mm, width=\textwidth,valign=b]{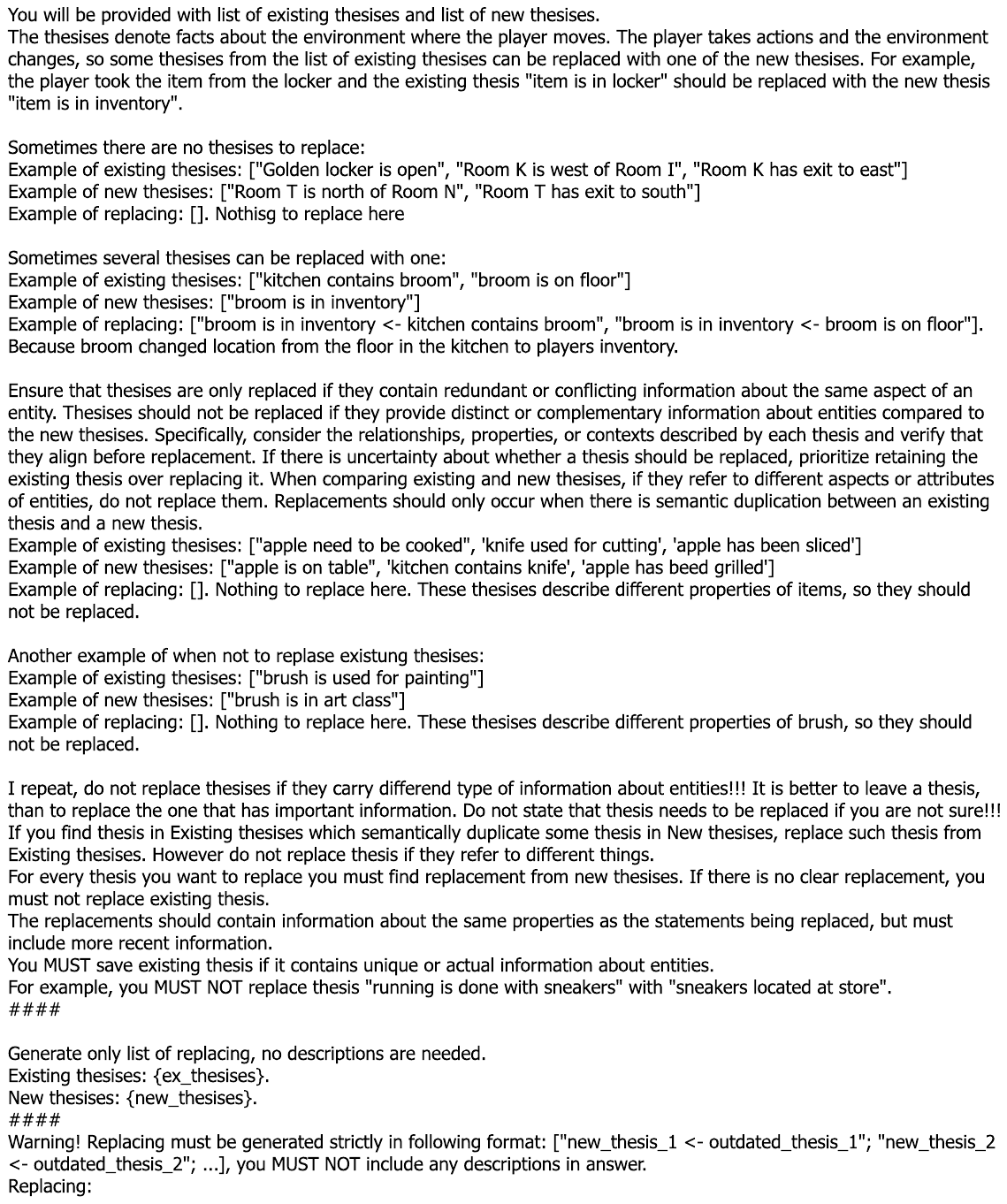}
			\end{minipage}
			\\ \hline

			\multirow{1}{*}{\textbf{Assistant}} &  -
			\\ \hline
		
		\end{tabular}%
	}
	\caption{LLM prompts for identifying outdated thesis memories}
	\label{tab:mempropmt_rthesis}
\end{table}

\begin{table}[H]
	\centering
	\resizebox{\textwidth}{!}{%
		\begin{tabular}{|c|l|}
			\hline
			\textbf{Type} & \multicolumn{1}{c|}{\textbf{Prompt}} \\ \hline
			\multirow{1}{*}{\textbf{System}} & 
			\begin{minipage}{\textwidth}
				\includegraphics[clip,trim={.07\textwidth} {\textheight} {.08\textwidth} 12mm, width=\textwidth,valign=b]{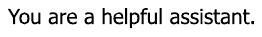}
			\end{minipage}
			\\ \hline
			
			\multirow{1}{*}{\textbf{User}} & 
			\begin{minipage}{\textwidth}
				\includegraphics[clip,trim={.07\textwidth} {.23\textheight} {.08\textwidth} 12mm, width=\textwidth,valign=b]{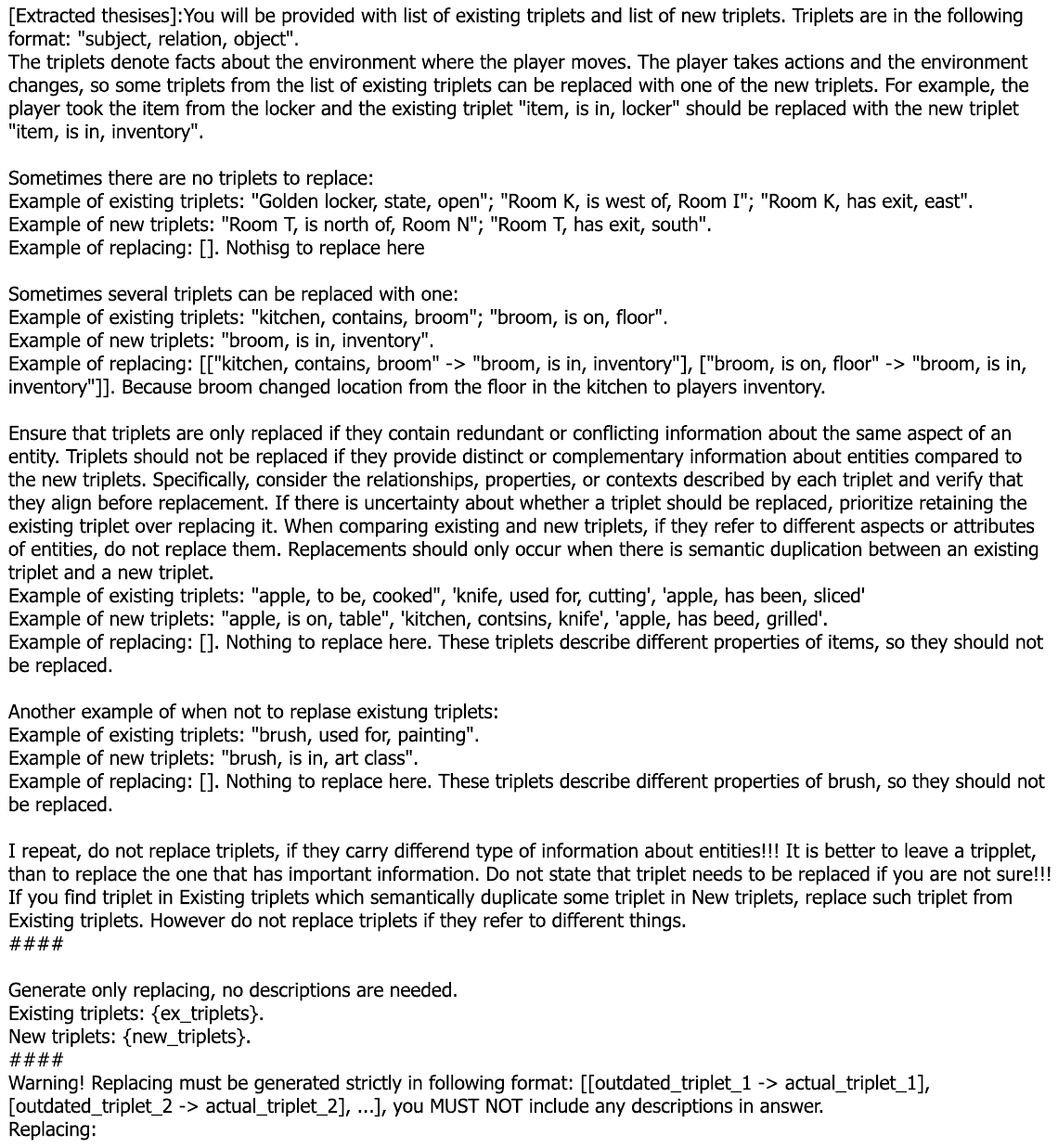}
			\end{minipage}
			\\ \hline
			
			\multirow{1}{*}{\textbf{Assistant}} &  -
			\\ \hline
			
		\end{tabular}%
	}
	\caption{LLM prompts for detecting obsolete object memories}
	\label{tab:mempropmt_rsimple}
\end{table}

\section{LLM prompts used in proposed QA pipeline}
\label{app:weak_qa_prompts}

Table \ref{tab:wqaprompt_kwe} presents the LLM prompts employed in the QA pipeline at the second stage for extracting key entities from the original user question. Table \ref{tab:wqaprompt_ag} displays the LLM prompts used in the fourth stage to generate contextually appropriate responses to the user question.

\begin{table}[H]
	\centering
	\resizebox{\textwidth}{!}{%
		\begin{tabular}{|c|l|}
			\hline
			\textbf{Type} & \multicolumn{1}{c|}{\textbf{Prompt}} \\ \hline
			\multirow{1}{*}{\textbf{System}} & 
			\begin{minipage}{\textwidth}
				\includegraphics[clip,trim={.07\textwidth} {.78\textheight} {.08\textwidth} 12mm, width=\textwidth,valign=b]{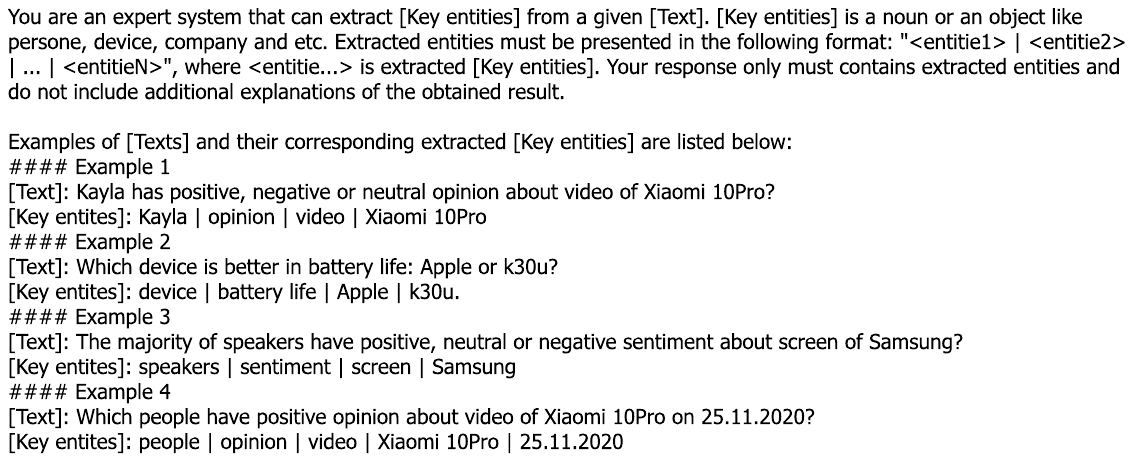}
			\end{minipage}
			\\ \hline
			
			\multirow{1}{*}{\textbf{User}} & 
			\begin{minipage}{\textwidth}
				\includegraphics[clip,trim={.07\textwidth} {\textheight} {.08\textwidth} 12mm, width=\textwidth,valign=b]{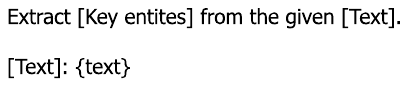}
			\end{minipage}
			\\ \hline
			
			\multirow{1}{*}{\textbf{Assistant}} & 
			\begin{minipage}{\textwidth}
				\includegraphics[clip,trim={.07\textwidth} {\textheight} {.08\textwidth} 12mm, width=\textwidth,valign=b]{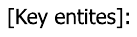}
			\end{minipage}
			\\ \hline
		\end{tabular}%
	}
	\caption{LLM prompts for extracting key entities from natural language text}
	\label{tab:wqaprompt_kwe}
\end{table}

\begin{table}[H]
	\centering
	\resizebox{\textwidth}{!}{%
		\begin{tabular}{|c|l|}
			\hline
			\textbf{Type} & \multicolumn{1}{c|}{\textbf{Prompt}} \\ \hline
			\multirow{1}{*}{\textbf{System}} & 
			\begin{minipage}{\textwidth}
				\includegraphics[clip,trim={.07\textwidth} {.31\textheight} {.08\textwidth} 12mm, width=\textwidth,valign=b]{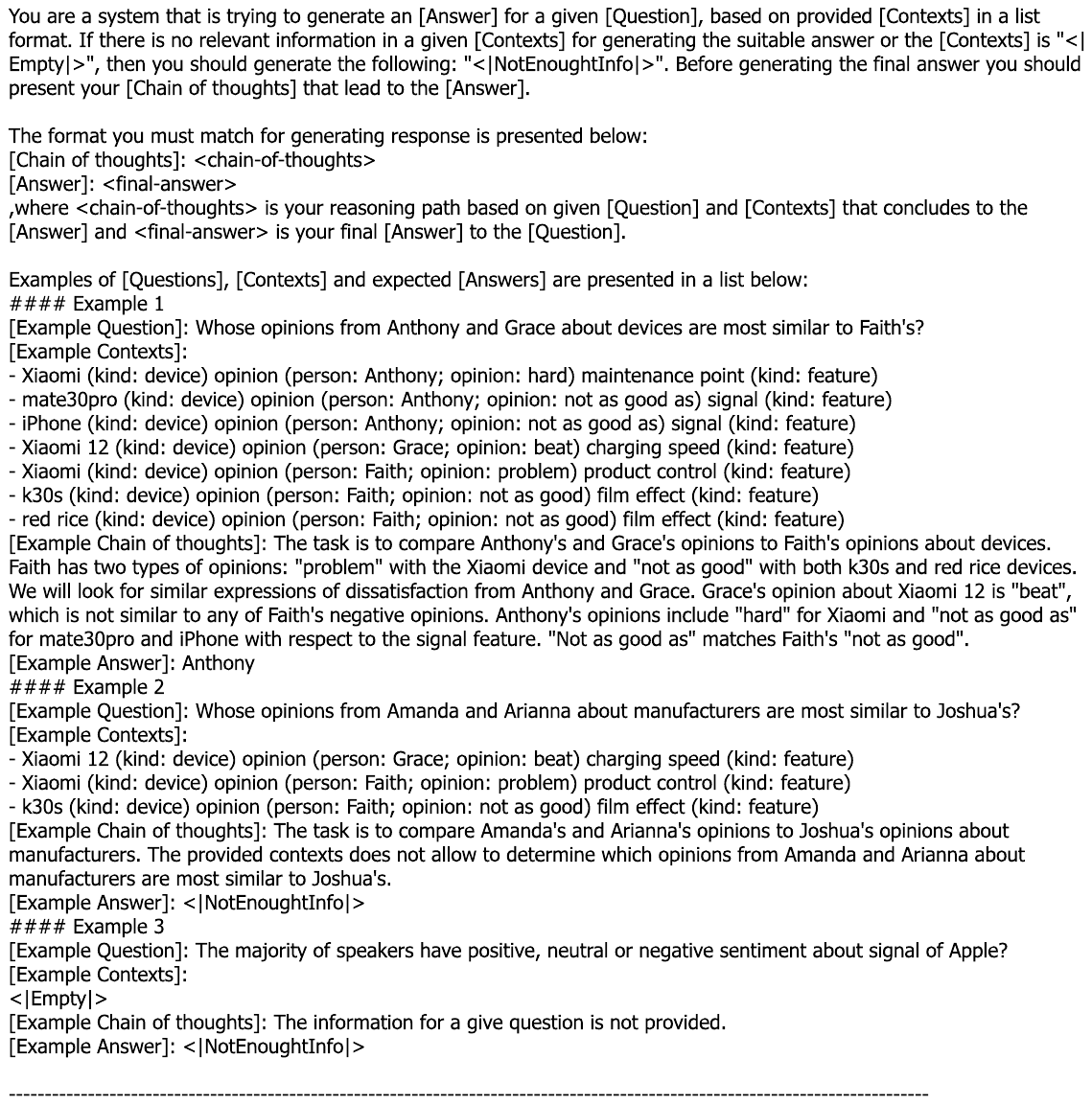}
			\end{minipage}
			\\ \hline
			
			\multirow{1}{*}{\textbf{User}} & 
			\begin{minipage}{\textwidth}
				\includegraphics[clip,trim={.07\textwidth} {\textheight} {.08\textwidth} 12mm, width=\textwidth,valign=b]{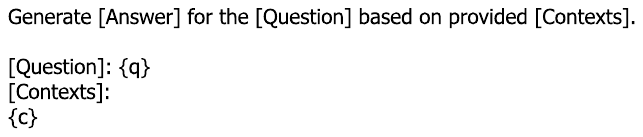}
			\end{minipage}
			\\ \hline
			
			\multirow{1}{*}{\textbf{Assistant}} & 
			\begin{minipage}{\textwidth}
				\includegraphics[clip,trim={.07\textwidth} {\textheight} {.08\textwidth} 12mm, width=\textwidth,valign=b]{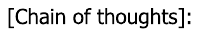}
			\end{minipage}
			\\ \hline
		\end{tabular}%
	}
\caption{LLM prompts for the conditional generation of an answer to the user question}
	\label{tab:wqaprompt_ag}
\end{table}

\section{Preprocessing operations for evaluation datasets}
\label{app:datasets}

For the original HotpotQA dataset, the distractor/validation subset was selected, comprising 7405 question-answer (QA) pairs and 13781 unique contexts. QA pairs were then filtered to exclude those with associated contexts falling outside a specified length range (measured in characters), retaining only contexts between 64 and 1024 characters in length. This filtering process resulted in 13291 remaining contexts. Finally, first 2000 QA pairs and their corresponding contexts were extracted, yielding a final subset of 3933 unique contexts.

For the original TriviaQA dataset, the rc.wikipedia/validation subset was selected, comprising 7993 question-answer (QA) pairs. Given the extensive length of the contexts in this dataset, they were segmented into smaller fragments (chunks) using the "RecursiveCharacterTextSplitter" class from the LangChain library. The following hyperparameters were applied: a chunk size of 1024 characters, separators set to double newline characters ("$\backslash$n$\backslash$n"), a chunk overlap of 64 characters, the "len"function for length calculation, and is\_separator\_regex set to "False". This preprocessing yielded 278384 unique text fragments. Subsequently, QA pairs were excluded if their associated text fragments fell outside the specified length bounds (minimum 64 and maximum 1024 characters), resulting in 13291 retained fragments. Additionally, since the original contexts were split without explicit tracking of which fragment contained the necessary information to answer a given question, if any fragment from a context was discarded, all remaining fragments from that context were also removed to ensure coherence. This step further reduced the dataset to 9975 unique fragments. Finally, the first 500 QA pairs and their corresponding relevant fragments were selected, leaving a total of 4925 unique fragments for analysis.  

Thus, evaluation sets for evaluation of proposed/implemented Memorize and QA pipelines were obtained. The characteristics of the obtained subsets of HotpotQA, TriviaQA and DiaASQ datasets can be found in Table \ref{tab:datasets_stats}.

\begin{table}[H]
	\centering
	\renewcommand{\arraystretch}{1.5}
	\resizebox{\textwidth}{!}{%
		\begin{tabular}{|c|ccccccc|cccc|}
			\hline
			\multirow{4}{*}{\textbf{Dataset}} & \multicolumn{7}{c|}{\textbf{QA-pairs}} & \multicolumn{4}{c|}{\textbf{Relevant contexts}} \\ \cline{2-12} 
			& \multicolumn{1}{c|}{\multirow{2}{*}{\textbf{Amount}}} & \multicolumn{3}{c|}{\begin{tabular}[c]{@{}c@{}}\textbf{Questions length} \\ \textbf{(in characters)}\end{tabular}} & \multicolumn{3}{c|}{\begin{tabular}[c]{@{}c@{}}\textbf{Answers length} \\ \textbf{(in characters)}\end{tabular}} & \multicolumn{1}{c|}{\multirow{2}{*}{\textbf{Amount}}} & \multicolumn{3}{c|}{\begin{tabular}[c]{@{}c@{}}\textbf{Length} \\ \textbf{(in characters)}\end{tabular}} \\ \cline{3-8} \cline{10-12} 
			& \multicolumn{1}{c|}{} & \multicolumn{1}{c|}{\textbf{median}} & \multicolumn{1}{c|}{\textbf{mean}} & \multicolumn{1}{c|}{\textbf{std}} & \multicolumn{1}{c|}{\textbf{median}} & \multicolumn{1}{c|}{\textbf{mean}} & \textbf{std} & \multicolumn{1}{c|}{} & \multicolumn{1}{c|}{\textbf{median}} & \multicolumn{1}{c|}{\textbf{mean}} & \textbf{std} \\ \hline
			\textbf{DiaASQ} & \multicolumn{1}{c|}{5698} & \multicolumn{1}{c|}{114} & \multicolumn{1}{c|}{109.44} & \multicolumn{1}{c|}{18.66} & \multicolumn{1}{c|}{8} & \multicolumn{1}{c|}{7.57} & 2.30 & \multicolumn{1}{c|}{3483} & \multicolumn{1}{c|}{556} & \multicolumn{1}{c|}{613.00} & 324.35 \\ \hline
			\textbf{HotpotQA} & \multicolumn{1}{c|}{2000} & \multicolumn{1}{c|}{87} & \multicolumn{1}{c|}{92.98} & \multicolumn{1}{c|}{32.62} & \multicolumn{1}{c|}{13} & \multicolumn{1}{c|}{15.29} & 11.87 & \multicolumn{1}{c|}{3933} & \multicolumn{1}{c|}{384} & \multicolumn{1}{c|}{413.72} & 201.05 \\ \hline
			\multicolumn{1}{|l|}{\textbf{TriviaQA}} & \multicolumn{1}{c|}{500} & \multicolumn{1}{c|}{66} & \multicolumn{1}{c|}{76.37} & \multicolumn{1}{c|}{39.46} & \multicolumn{1}{c|}{9} & \multicolumn{1}{c|}{10.17} & 5.76 & \multicolumn{1}{c|}{4925} & \multicolumn{1}{c|}{807} & \multicolumn{1}{c|}{765.11} & 196.32 \\ \hline
		\end{tabular}}
	\renewcommand{\arraystretch}{1}
	\caption{Extended characteristics of datasets, used to evaluate proposed Memorize and QA pipelines}
	\label{tab:datasets_stats}
\end{table}

\section{Retrieval hyperparameters}
\label{app:retriev_hyper}
\begin{itemize}
\item \textbf{A*}: h\_metric\_name -- ip; max\_depth -- 10; max\_passed\_nodes -- 150.
\item \textbf{WaterCircles}: strict\_filter -- True; hyper\_num -- 15; episodic\_num -- 15; chain\_triplets\_num -- 25; other\_triplets\_num -- 6; do\_text\_pruning -- False. 
\item \textbf{BeamSearch}: max\_depth -- 5; max\_paths -- 10; same\_path\_intersection\_by\_node -- False; diff\_paths\_intersection\_by\_node -- False; diff\_paths\_intersection\_by\_rel -- False; mean\_alpha -- 0.75; final\_sorting\_mode -- mixed.
\end{itemize}

\section{LLM--as--a--Judge instructions}
\label{app:judgescore_hyperp}

To ensure the reproducibility of the obtained results, LLM inference was conducted using a deterministic generation strategy. The following hyperparameters were applied: num\_predict -- 2048, seed -- 42, temperature -- 0.0, and top\_k -- 1. The Qwen2.5 7B model, sourced from the Ollama repository, was prompted to evaluate whether the outputs of the proposed QA pipeline correctly answered the given questions. The specific LLM prompts used for this assessment are provided in Table \ref{tab:prompts_judge}.

\begin{table}[H]
	\centering
	\resizebox{\textwidth}{!}{%
		\begin{tabular}{|c|l|}
			\hline
			\textbf{Type} & \multicolumn{1}{c|}{\textbf{Prompt}} \\ \hline
			\multirow{1}{*}{\textbf{System}} & 
			\begin{minipage}{\textwidth}
				\includegraphics[clip,trim={.07\textwidth} {.55\textheight} {.08\textwidth} 12mm, width=\textwidth,valign=b]{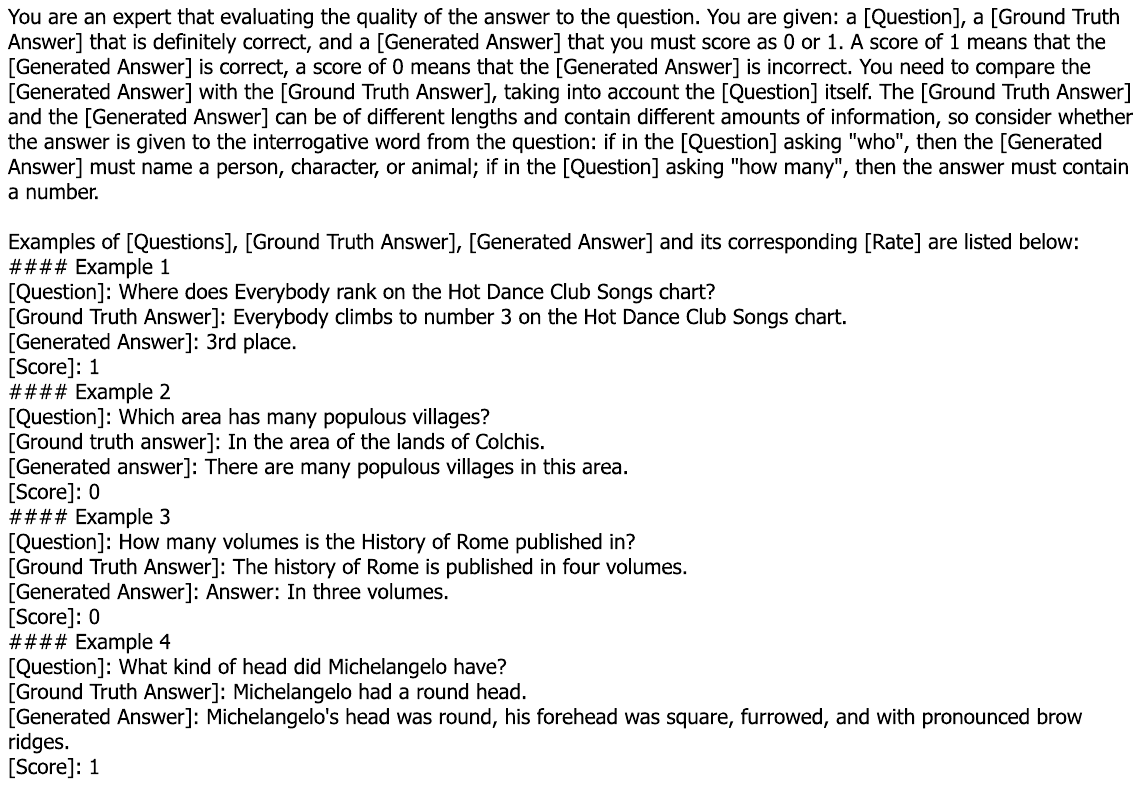}
			\end{minipage}
			\\ \hline
			
			\multirow{1}{*}{\textbf{User}} & 
			\begin{minipage}{\textwidth}
				\includegraphics[clip,trim={.07\textwidth} {\textheight} {.08\textwidth} 12mm, width=\textwidth,valign=b]{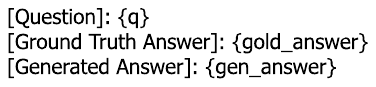}
			\end{minipage}
			\\ \hline
			
			\multirow{1}{*}{\textbf{Assistant}} & 
			\begin{minipage}{\textwidth}
				\includegraphics[clip,trim={.07\textwidth} {\textheight} {.08\textwidth} 12mm, width=\textwidth,valign=b]{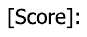}
			\end{minipage}
			\\ \hline
		\end{tabular}%
	}
	\caption{LLM prompts used in LLM--as--a--Judge framework}
	\label{tab:prompts_judge}
\end{table}

\section{Characteristics of constructed memory graphs}
\label{app:constr_graphs}

To evaluate the QA pipeline we constructed 14 memory graphs based on the given dataset and LLM configurations. The structural characteristics of the generated graphs are detailed in Table \ref{tbl:kg_stats}. 

Results indicate that certain graphs experienced parsing errors during the extraction of LLM responses, resulting in incomplete storage of contextual information. Across the evaluated datasets, the average parsing error rates were as follows: DiaASQ (7.0\%), HotpotQA (6.3\%), and TriviaQA (7.3\%). For the DiaASQ dataset, DeepSeek R1 7B produced the highest number of thesis and object vertices, as well as hyper-edges, while GPT-4o-mini extracted the largest number of unique object-typed edges. In contrast, for the HotpotQA and TriviaQA datasets, Qwen2.5 7B generated the most thesis and object vertices, whereas DeepSeek R1 7B again yielded the highest number of hyper-edges. Based on the obtained results, we can model the structural characteristics of memory graphs generated by each evaluated system; see Figure \ref{tab:meankg_stats}.

The data reveal significant variation in LLM response parsing accuracy across models. Specifically, Qwen2.5 7B and Llama3.1 8B demonstrate lowest error rates (0.02\% each), while DeepSeek V3 exhibits the highest parsing error rate (31.21\%). Intermediate performance is observed in DeepSeekR1 7B (0.29\%) and GPT4o mini (9.87\%). Regarding memory graph composition, DeepSeekR1 7B and Qwen2.5 7B yield the most comprehensive representations, generating the highest number of thesis/object memories and associated edges. Further analysis of vertices creation efficiency per contextual unit shows Qwen2.5 7B achieves superior granularity, producing the largest number of thesis/object memories while maintaining contextual coherence: see Table \ref{tab:kfinfo_perepisodic}.  

\begin{table}[H]
	\centering
	\renewcommand{\arraystretch}{1.8}
		\resizebox{0.7\textwidth}{!}{%
		\makebox[\textwidth]{%
			\rule{0cm}{1mm}
		\begin{tabular}{c|ccccccccccc|}
			\hline
			\multicolumn{1}{|c|}{\multirow{3}{*}{\textbf{LLM}}} & \multicolumn{1}{c|}{\multirow{3}{*}{\begin{tabular}[c]{@{}c@{}}\textbf{Number of contexts} \\ \textbf{to store in graph}\end{tabular}}} & \multicolumn{3}{c|}{\textbf{Nubmer of vertices}} & \multicolumn{3}{c|}{\textbf{Number of edges}} & \multicolumn{4}{c|}{\textbf{Average number of matching vertices (by type)}} \\ \cline{3-12} 
			\multicolumn{1}{|c|}{} & \multicolumn{1}{c|}{} & \multicolumn{1}{c|}{\textbf{episodic}} & \multicolumn{1}{c|}{\textbf{thesis}} & \multicolumn{1}{c|}{\textbf{object}} & \multicolumn{1}{c|}{\textbf{episodic}} & \multicolumn{1}{c|}{\textbf{thesis}} & \multicolumn{1}{c|}{\textbf{object}} & \multicolumn{1}{c|}{\begin{tabular}[c]{@{}c@{}}\textbf{object neighbours} \\ \textbf{(to episodic vertices)}\end{tabular}} & \multicolumn{1}{c|}{\begin{tabular}[c]{@{}c@{}}\textbf{thesis neighbours} \\ \textbf{(to episodic vertices)}\end{tabular}} & \multicolumn{1}{c|}{\begin{tabular}[c]{@{}c@{}}\textbf{object neighbours} \\ \textbf{(to thesis vertices)}\end{tabular}} & \begin{tabular}[c]{@{}c@{}}\textbf{object neighbours} \\ \textbf{(to object vertices)}\end{tabular} \\ \hline
			& \multicolumn{11}{c|}{\textbf{DiaASQ}} \\ \hline
			\multicolumn{1}{|c|}{\textbf{DeepSeek R1 7B}} & \multicolumn{1}{c|}{\multirow{5}{*}{3483}} & \multicolumn{1}{c|}{3477} & \multicolumn{1}{c|}{\underline{34039}} & \multicolumn{1}{c|}{\underline{30097}} & \multicolumn{1}{c|}{\underline{138584}} & \multicolumn{1}{c|}{\underline{133301}} & \multicolumn{1}{c|}{34049} & \multicolumn{1}{c|}{\underline{30.06}} & \multicolumn{1}{c|}{\underline{10.01}} & \multicolumn{1}{c|}{\underline{3.91}} & 1.99 \\ \cline{1-1} \cline{3-12} 
			\multicolumn{1}{|c|}{\textbf{Qwen2.5 7B}} & \multicolumn{1}{c|}{} & \multicolumn{1}{c|}{\underline{3482}} & \multicolumn{1}{c|}{32512} & \multicolumn{1}{c|}{28420} & \multicolumn{1}{c|}{129974} & \multicolumn{1}{c|}{111618} & \multicolumn{1}{c|}{31290} & \multicolumn{1}{c|}{27.98} & \multicolumn{1}{c|}{9.48} & \multicolumn{1}{c|}{3.43} & 2.04 \\ \cline{1-1} \cline{3-12} 
			\multicolumn{1}{|c|}{\textbf{Llama3.1 8B}} & \multicolumn{1}{c|}{} & \multicolumn{1}{c|}{\underline{3482}} & \multicolumn{1}{c|}{29655} & \multicolumn{1}{c|}{20014} & \multicolumn{1}{c|}{100045} & \multicolumn{1}{c|}{77467} & \multicolumn{1}{c|}{28063} & \multicolumn{1}{c|}{20.21} & \multicolumn{1}{c|}{8.52} & \multicolumn{1}{c|}{2.61} & \underline{2.25} \\ \cline{1-1} \cline{3-12} 
			\multicolumn{1}{|c|}{\textbf{GPT4o mini}} & \multicolumn{1}{c|}{} & \multicolumn{1}{c|}{\underline{3482}} & \multicolumn{1}{c|}{31361} & \multicolumn{1}{c|}{28172} & \multicolumn{1}{c|}{125403} & \multicolumn{1}{c|}{96667} & \multicolumn{1}{c|}{\underline{39477}} & \multicolumn{1}{c|}{27.00} & \multicolumn{1}{c|}{9.00} & \multicolumn{1}{c|}{3.08} & 1.89 \\ \cline{1-1} \cline{3-12} 
			\multicolumn{1}{|c|}{\textbf{DeepSeek V3}} & \multicolumn{1}{c|}{} & \multicolumn{1}{c|}{2270} & \multicolumn{1}{c|}{21482} & \multicolumn{1}{c|}{18416} & \multicolumn{1}{c|}{80118} & \multicolumn{1}{c|}{64349} & \multicolumn{1}{c|}{28877} & \multicolumn{1}{c|}{25.83} & \multicolumn{1}{c|}{9,46} & \multicolumn{1}{c|}{2.99} & 2.11 \\ \hline
			& \multicolumn{11}{c|}{\textbf{HotpotQA}} \\ \hline
			\multicolumn{1}{|c|}{\textbf{DeepSeek R1 7B}} & \multicolumn{1}{c|}{\multirow{5}{*}{3933}} & \multicolumn{1}{c|}{3921} & \multicolumn{1}{c|}{27137} & \multicolumn{1}{c|}{55254} & \multicolumn{1}{c|}{\underline{121359}} & \multicolumn{1}{c|}{\underline{111174}} & \multicolumn{1}{c|}{33714} & \multicolumn{1}{c|}{\underline{24.00}} & \multicolumn{1}{c|}{7.34} & \multicolumn{1}{c|}{\underline{4.09}} & \underline{1.55} \\ \cline{1-1} \cline{3-12} 
			\multicolumn{1}{|c|}{\textbf{Qwen2.5 7B}} & \multicolumn{1}{c|}{} & \multicolumn{1}{c|}{\underline{3933}} & \multicolumn{1}{c|}{\underline{31795}} & \multicolumn{1}{c|}{\underline{56078}} & \multicolumn{1}{c|}{119387} & \multicolumn{1}{c|}{106037} & \multicolumn{1}{c|}{38460} & \multicolumn{1}{c|}{22.26} & \multicolumn{1}{c|}{\underline{8.15}} & \multicolumn{1}{c|}{3.33} & 1.36 \\ \cline{1-1} \cline{3-12} 
			\multicolumn{1}{|c|}{\textbf{Llama3.1 8B}} & \multicolumn{1}{c|}{} & \multicolumn{1}{c|}{\underline{3933}} & \multicolumn{1}{c|}{26364} & \multicolumn{1}{c|}{40601} & \multicolumn{1}{c|}{89506} & \multicolumn{1}{c|}{75332} & \multicolumn{1}{c|}{29021} & \multicolumn{1}{c|}{16.04} & \multicolumn{1}{c|}{6.79} & \multicolumn{1}{c|}{2.85} & 1.47 \\ \cline{1-1} \cline{3-12} 
			\multicolumn{1}{|c|}{\textbf{GPT4o mini}} & \multicolumn{1}{c|}{} & \multicolumn{1}{c|}{\underline{3933}} & \multicolumn{1}{c|}{30777} & \multicolumn{1}{c|}{48771} & \multicolumn{1}{c|}{105524} & \multicolumn{1}{c|}{93791} & \multicolumn{1}{c|}{\underline{42599}} & \multicolumn{1}{c|}{18.99} & \multicolumn{1}{c|}{7.94} & \multicolumn{1}{c|}{3.04} & 1.36 \\ \cline{1-1} \cline{3-12} 
			\multicolumn{1}{|c|}{\textbf{DeepSeek V3}} & \multicolumn{1}{c|}{} & \multicolumn{1}{c|}{2713} & \multicolumn{1}{c|}{20164} & \multicolumn{1}{c|}{35291} & \multicolumn{1}{c|}{73242} & \multicolumn{1}{c|}{63921} & \multicolumn{1}{c|}{34621} & \multicolumn{1}{c|}{19.55} & \multicolumn{1}{c|}{7.53} & \multicolumn{1}{c|}{3.17} & 1.41 \\ \hline
			& \multicolumn{11}{c|}{\textbf{TriviaQA}} \\ \hline
			\multicolumn{1}{|c|}{\textbf{DeepSeek R1 7B}} & \multicolumn{1}{c|}{\multirow{4}{*}{4925}} & \multicolumn{1}{c|}{4905} & \multicolumn{1}{c|}{48855} & \multicolumn{1}{c|}{96132} & \multicolumn{1}{c|}{213481} & \multicolumn{1}{c|}{\underline{201861}} & \multicolumn{1}{c|}{52019} & \multicolumn{1}{c|}{33.50} & \multicolumn{1}{c|}{10.56} & \multicolumn{1}{c|}{\underline{4.13}} & 1.53 \\ \cline{1-1} \cline{3-12} 
			\multicolumn{1}{|c|}{\textbf{Qwen2.5 7B}} & \multicolumn{1}{c|}{} & \multicolumn{1}{c|}{\underline{4923}} & \multicolumn{1}{c|}{\underline{52835}} & \multicolumn{1}{c|}{\underline{109900}} & \multicolumn{1}{c|}{\underline{220991}} & \multicolumn{1}{c|}{188780} & \multicolumn{1}{c|}{\underline{62043}} & \multicolumn{1}{c|}{\underline{34.15}} & \multicolumn{1}{c|}{\underline{10.87}} & \multicolumn{1}{c|}{3.57} & 1.27 \\ \cline{1-1} \cline{3-12} 
			\multicolumn{1}{|c|}{\textbf{Llama3.1 8B}} & \multicolumn{1}{c|}{} & \multicolumn{1}{c|}{4922} & \multicolumn{1}{c|}{45241} & \multicolumn{1}{c|}{72285} & \multicolumn{1}{c|}{158202} & \multicolumn{1}{c|}{127389} & \multicolumn{1}{c|}{46757} & \multicolumn{1}{c|}{22.83} & \multicolumn{1}{c|}{9.35} & \multicolumn{1}{c|}{2.81} & \underline{1.54} \\ \cline{1-1} \cline{3-12} 
			\multicolumn{1}{|c|}{\textbf{DeepSeek V3}} & \multicolumn{1}{c|}{} & \multicolumn{1}{c|}{3506} & \multicolumn{1}{c|}{37496} & \multicolumn{1}{c|}{68602} & \multicolumn{1}{c|}{143933} & \multicolumn{1}{c|}{122480} & \multicolumn{1}{c|}{61821} & \multicolumn{1}{c|}{30.32} & \multicolumn{1}{c|}{10.75} & \multicolumn{1}{c|}{3.26} & 1.42 \\ \hline
		\end{tabular}%
	}}
	\renewcommand{\arraystretch}{1}
	\caption{Characteristics of constructed memory graphs for QA experiments}
	\label{tbl:kg_stats}
\end{table}

\begin{table}[H]
	\centering
	\renewcommand{\arraystretch}{1.7}
		\resizebox{0.71\textwidth}{!}{%
	\makebox[\textwidth]{%
		\rule{0cm}{1mm}
		\begin{tabular}{|c|c|ccc|ccc|cccc|}
			\hline
			\multirow{3}{*}{\textbf{LLM}} & \multirow{3}{*}{\begin{tabular}[c]{@{}c@{}}\textbf{Number of contexts} \\ \textbf{to store in graph}\end{tabular}} & \multicolumn{3}{c|}{\textbf{Number of vertices}} & \multicolumn{3}{c|}{\textbf{Number of edges}} & \multicolumn{4}{c|}{\textbf{Average number of adjacent vertices (by type)}} \\ \cline{3-12} 
			&  & \multicolumn{1}{c|}{\multirow{2}{*}{\textbf{episodic}}} & \multicolumn{1}{c|}{\multirow{2}{*}{\textbf{thesis}}} & \multirow{2}{*}{\textbf{object}} & \multicolumn{1}{c|}{\multirow{2}{*}{\textbf{episodic}}} & \multicolumn{1}{c|}{\multirow{2}{*}{\textbf{thesis}}} & \multirow{2}{*}{\textbf{object}} & \multicolumn{1}{c|}{\multirow{2}{*}{\begin{tabular}[c]{@{}c@{}}\textbf{object neighbours} \\ \textbf{(to episodic vertices)}\end{tabular}}} & \multicolumn{1}{c|}{\multirow{2}{*}{\begin{tabular}[c]{@{}c@{}}\textbf{thesis neighbours} \\ \textbf{(to episodic vertices)}\end{tabular}}} & \multicolumn{1}{c|}{\multirow{2}{*}{\begin{tabular}[c]{@{}c@{}}\textbf{object neighbours} \\ \textbf{(to thesis vertices)}\end{tabular}}} & \multirow{2}{*}{\begin{tabular}[c]{@{}c@{}}\textbf{object neighbours} \\ \textbf{(to object vertices)}\end{tabular}} \\
			&  & \multicolumn{1}{c|}{} & \multicolumn{1}{c|}{} &  & \multicolumn{1}{c|}{} & \multicolumn{1}{c|}{} &  & \multicolumn{1}{c|}{} & \multicolumn{1}{c|}{} & \multicolumn{1}{c|}{} &  \\ \hline
			\textbf{DeepSeek R1 7B} & \multirow{5}{*}{4113} & \multicolumn{1}{c|}{4101} & \multicolumn{1}{c|}{36677} & 60494 & \multicolumn{1}{c|}{\underline{157808}} & \multicolumn{1}{c|}{\underline{148778}} & 39927 & \multicolumn{1}{c|}{\underline{29.18}} & \multicolumn{1}{c|}{9.30} & \multicolumn{1}{c|}{\underline{4.04}} & 1.69 \\ \cline{1-1} \cline{3-12} 
			\textbf{Qwen2.5 7B} &  & \multicolumn{1}{c|}{\underline{4112}} & \multicolumn{1}{c|}{\underline{39047}} & \underline{64799} & \multicolumn{1}{c|}{156784} & \multicolumn{1}{c|}{135478} & \underline{43931} & \multicolumn{1}{c|}{28.13} & \multicolumn{1}{c|}{\underline{9.5}} & \multicolumn{1}{c|}{3.44} & 1.55 \\ \cline{1-1} \cline{3-12} 
			\textbf{Llama3.1 8B} &  & \multicolumn{1}{c|}{\underline{4112}} & \multicolumn{1}{c|}{33753} & 44300 & \multicolumn{1}{c|}{115917} & \multicolumn{1}{c|}{93396} & 34613 & \multicolumn{1}{c|}{19.69} & \multicolumn{1}{c|}{8.22} & \multicolumn{1}{c|}{2.75} & \underline{1.75} \\ \cline{1-1} \cline{3-12} 
			\textbf{GPT4o mini} &  & \multicolumn{1}{c|}{3707} & \multicolumn{1}{c|}{31069} & 38471 & \multicolumn{1}{c|}{115463} & \multicolumn{1}{c|}{95229} & 41038 & \multicolumn{1}{c|}{22.99} & \multicolumn{1}{c|}{8.47} & \multicolumn{1}{c|}{3.06} & 1.625 \\ \cline{1-1} \cline{3-12} 
			\textbf{DeepSeek V3} &  & \multicolumn{1}{c|}{2829} & \multicolumn{1}{c|}{26380} & 40769 & \multicolumn{1}{c|}{99097} & \multicolumn{1}{c|}{83583} & 41773 & \multicolumn{1}{c|}{25.23} & \multicolumn{1}{c|}{9.24} & \multicolumn{1}{c|}{3.14} & 1.64 \\ \hline
		\end{tabular}%
	}}
	\renewcommand{\arraystretch}{1}
	\caption{Average/expected characteristics of knowledge graphs in case of using given LLM models for their construction}
	\label{tab:meankg_stats}
\end{table}

\begin{table}[H]
	\centering
	\renewcommand{\arraystretch}{1.2}
	\resizebox{0.4\textwidth}{!}{%
		\begin{tabular}{|c|cc|cc|}
			\hline
			\multirow{2}{*}{\textbf{LLM}} & \multicolumn{2}{c|}{\textbf{Vertex}} & \multicolumn{2}{c|}{\textbf{Edge}} \\ \cline{2-5} 
			& \multicolumn{1}{c|}{\textbf{thesis}} & \multicolumn{1}{c|}{\textbf{object}} & \multicolumn{1}{c|}{\textbf{thesis}} & \textbf{object} \\ \hline
			\textbf{DeepSeek R1 7B} & \multicolumn{1}{c|}{9} & \multicolumn{1}{c|}{15} & \multicolumn{1}{c|}{\underline{36}} & 10 \\ \hline
			\textbf{Qwen2.5 7B} & \multicolumn{1}{c|}{\underline{10}} & \multicolumn{1}{c|}{\underline{16}} & \multicolumn{1}{c|}{33} & 11 \\ \hline
			\textbf{Llama3.1 8B} & \multicolumn{1}{c|}{8} & \multicolumn{1}{c|}{11} & \multicolumn{1}{c|}{23} & 8 \\ \hline
			\textbf{GPT4o mini} & \multicolumn{1}{c|}{8} & \multicolumn{1}{c|}{10} & \multicolumn{1}{c|}{26} & 11 \\ \hline
			\textbf{DeepSeek V3} & \multicolumn{1}{c|}{9} & \multicolumn{1}{c|}{14} & \multicolumn{1}{c|}{30} & \underline{15} \\ \hline
		\end{tabular}%
	}
	\renewcommand{\arraystretch}{1}
	\caption{Number of unique vertices/edges that were added to the graph when processing (using a given LLM model) and storing one episodic memory (episodic verrtex)}
	\label{tab:kfinfo_perepisodic}
\end{table}

In addition to characteristics of constructed graphs, we collected information about time and speed of the Memorize pipeline, which responsible for parsing incoming unstructured texts in natural language and storing them in memory model: see Table \ref{tab:gconstr-time}.

\begin{table}[H]
\centering
\renewcommand{\arraystretch}{1.5}
\resizebox{\textwidth}{!}{%
\begin{tabular}{|c|ccccc|c}
\hline
\multirow{2}{*}{\begin{tabular}[c]{@{}c@{}} \textbf{LLM} / \\ \textbf{Dataset}\end{tabular}} & \multicolumn{5}{c|}{\textbf{Memory construction time (hours) and speed (contexts per minute)}} & \multicolumn{1}{c|}{\multirow{2}{*}{Mean:}} \\ \cline{2-6}
  & \multicolumn{1}{c|}{\textbf{Qwen2.5 7B}} & \multicolumn{1}{c|}{\textbf{DeepSeek R1 7B}} & \multicolumn{1}{c|}{\textbf{Llama3.1 8B}} & \multicolumn{1}{c|}{\textbf{GPT4o mini}} & \textbf{DeepSeek V3} & \multicolumn{1}{c|}{} \\ \hline
\textbf{DiaASQ} & \multicolumn{1}{c|}{23.5 / 2.47} & \multicolumn{1}{c|}{57.5 / 1.00} & \multicolumn{1}{c|}{32 / 1.81} & \multicolumn{1}{c|}{32 / 1.81} & 68 / 0.85 & \multicolumn{1}{c|}{42.6 / 1.58} \\ \hline
\textbf{HotpotQA} & \multicolumn{1}{c|}{61.5 / 1.06} & \multicolumn{1}{c|}{47.5 / 1.38} & \multicolumn{1}{c|}{27 / 2.42} & \multicolumn{1}{c|}{34 / 1.92} & 72 / 0.91 & \multicolumn{1}{c|}{48.4 / 1.53} \\ \hline
\textbf{TriviaQA} & \multicolumn{1}{c|}{90 / 0.91} & \multicolumn{1}{c|}{-} & \multicolumn{1}{c|}{90 / 0.91} & \multicolumn{1}{c|}{-} & 80 / 1.02 & \multicolumn{1}{c|}{86.6 / 0.94} \\ \hline
Mean: & \multicolumn{1}{c|}{58.3 / 1.48} & \multicolumn{1}{c|}{52.5 / 1.19} & \multicolumn{1}{c|}{\underline{49.7 / 1.71}} & \multicolumn{1}{c|}{\underline{33 / 1.86}} & 73.3 / 0.92 &  \\ \cline{1-6}
\end{tabular}}
\caption{Time and speed of memory graphs construction based on a given dataset and LLM, which were used in proposed QA pipeline evaluation}
\label{tab:gconstr-time}
\end{table}

Table \ref{tab:gconstr-time} shows that highest speed of parsing and saving information into memory is observed in configurations of Memorize pipeline there GPT4o mini is set as LLM (1.86); in second place is LLama3.1 8B, that was deployed in Ollama Docker container on local machine.

It is also important to note the amount of disk space that constructed memory is required. When using Milvus as a database for storing vector representations of text, on selected datasets, one memory graph in our implementation occupies, approximately, 80-90 GB. If Qdrant is used as a vector storage, the same graphs will occupy, approximately, 4-6 GB each.

\section{Comparison with existing RAG and GraphRAG methods}
\label{app:rag_graphrag_cmpr}

Based on experimental results, a comparative table, summarizing best-performing QA configurations by Exact Match metric, was compiled (see Table \ref{tab:qabest_em_retriever}).

\begin{table}[H]
\centering
\renewcommand{\arraystretch}{1.5}
\resizebox{\textwidth}{!}{%
\begin{tabular}{|c|c|c|c|c|c|}
\hline
\begin{tabular}[c]{@{}c@{}} \textbf{LLM} / \\ \textbf{Dataset}\end{tabular} & \textbf{Qwen2.5 7B} & \textbf{DeepSeek R1 7B} & \textbf{Llama3.1 8B} & \textbf{GPT4o mini} & \textbf{DeepSeek V3} \\ \hline
\textbf{DiaASQ} & 0.22 / BS + WC / T & 0.11 / AS / E & 0.18 / AS / E & 0.47 / BS + WC / E & 0.46 / BS + WC / O \\ \hline
\textbf{HotpotQA} & 0.18 / AS / all & 0.14 / AS / E & 0.37 / BS / O & 0.59 / BS + WC / all & 0.6 / BS + WC / T \\ \hline
\textbf{TriviaQA} & 0.2 / BS / E & 0.18 / AS / E & 0.47 / BS / E & - & 0.62 / BS + WC / all \\ \hline
Mean: & 0.2 & 0.14 & 0.34 & 0.53 & \underline{0.56} \\ \hline
\end{tabular}}
\caption{\textbf{Best QA configurations ranked by the Exact Match metric across all experiments.} The corresponding cells contain the Exact Match score, the retrieval algorithm used, and the type of restriction applied to the memory graph during traversal. Shortcuts for retrieval algorithms: \textbf{BS} -- BeamSearch; \textbf{AS} -- A*; \textbf{BS + AS} -- hybrid of BeamSearch and A*; \textbf{BS + WC} -- hybrid of BeamSearch and WaterCircles. Shortcuts for graph restrictions: \textbf{all} -- no restrictions applied; \textbf{E} -- episodic vertices excluded from traversal; \textbf{T} -- thesis vertices excluded; \textbf{O} -- object vertices excluded.}
\label{tab:qabest_em_retriever}
\end{table}

As shown in Table \ref{tab:qabest_em_retriever}, Qwen2.5 achieved the best performance (0.2) among the 7B models. Among all evaluated configurations, the highest overall effectiveness (0.56) was reached by setups incorporating DeepSeek V3. Notably, the top-performing 7B configurations predominantly relied on A*, especially under the constraint that traversal through episodic vertices was restricted. In contrast, the best DeepSeek V3 configurations frequently adopted a hybrid strategy combining BeamSearch and WaterCircles. Across high-performing configurations more broadly, BeamSearch consistently appeared as a key component of the retrieval pipeline.

To evaluate the performance of optimal configurations derived from our framework against existing Retrieval-Augmented Generation (RAG) and GraphRAG approaches, we conducted a systematic literature review. The search was performed across five academic search engines: SciSpace, Scite, PaperDigest, Consensus, and Elicit. Using each engine, we selected the first ten publications in the search results for the following queries: in the case of RAG methods, "Retrieval-augmented generation methods based on pretrained language models" and "RAG methods in NLP"; in the case of GraphRAG methods, "RAG on Knowledge Graphs", "Enhancing RAG--approach with Knowledge Graphs", "Graph RAG", and "RAG with integration of Large Language Models (LLMs) and Knowledge Graphs (KGs)". The search was restricted to publications up to 2018. Subsequently, we applied a three-stage filtering process: (1) duplicate entries were removed, and papers introducing novel evaluation datasets were excluded; (2) works focusing on domain-specific applications without broader methodological contributions were discarded; (3) only studies providing comprehensive methodological descriptions and employing standardized benchmarks were retained. This process yielded two final sets of nine articles each, covering RAG and GraphRAG techniques, respectively. A comparative analysis of these methods against our framework's optimal graph construction and retrieval configuration is presented in Table \ref{tab:qamethods_compare}, with performance measured using the Exact Match metric.

\begin{table}[H]
\centering
\renewcommand{\arraystretch}{1.5}
\resizebox{\textwidth}{!}{%
\begin{tabular}{c|ccccccccc|c}
\hline
\multicolumn{1}{|c|}{\multirow{2}{*}{\textbf{Dataset}}} & \multicolumn{9}{c|}{\textbf{RAG method}} & \multicolumn{1}{c|}{\multirow{2}{*}{\textbf{Our method}}} \\ \cline{2-10}
\multicolumn{1}{|c|}{} & \multicolumn{1}{c|}{\textbf{REALM}~\cite{10.5555/3524938.3525306}} & \multicolumn{1}{c|}{\textbf{DPR}~\cite{karpukhin-etal-2020-dense}} & \multicolumn{1}{c|}{\textbf{RAG}~\cite{lewis2021retrievalaugmentedgenerationknowledgeintensivenlp}} & \multicolumn{1}{c|}{\textbf{ColBERT--QA}~\cite{khattab-etal-2021-relevance}} & \multicolumn{1}{c|}{\textbf{FiD}~\cite{izacard-grave-2021-leveraging}} & \multicolumn{1}{c|}{\textbf{EMDR2}~\cite{sachan2021endtoend}} & \multicolumn{1}{c|}{\textbf{RETRO}~\cite{Borgeaud2021ImprovingLM}} & \multicolumn{1}{c|}{\textbf{Atlas}~\cite{10.5555/3648699.3648950}} & \textbf{RePLUG}~\cite{shi-etal-2024-replug} & \multicolumn{1}{c|}{} \\ \hline
\multicolumn{1}{|c|}{\textbf{TriviaQA}} & \multicolumn{1}{c|}{53.9} & \multicolumn{1}{c|}{56.8} & \multicolumn{1}{c|}{55.8} & \multicolumn{1}{c|}{70.1} & \multicolumn{1}{c|}{67.6} & \multicolumn{1}{c|}{71.4} & \multicolumn{1}{c|}{62.1} & \multicolumn{1}{c|}{\underline{79.8}} & 77.3 & \multicolumn{1}{c|}{62.0} \\ \hline
 & \multicolumn{9}{c|}{\textbf{GraphRAG method}} &  \\ \cline{2-10}
 & \multicolumn{1}{c|}{\textbf{ToG}~\cite{sun2023thinkongraph}} & \multicolumn{1}{c|}{\textbf{RoG}~\cite{luo2024rog}} & \multicolumn{1}{c|}{\textbf{PMKGE}~\cite{liu2025enhancinglargelanguagemodels}} & \multicolumn{1}{c|}{\textbf{GRAG}~\cite{hu2024grag}} & \multicolumn{1}{c|}{\textbf{GNN--RAG}~\cite{mavromatis-karypis-2025-gnn}} & \multicolumn{1}{c|}{\textbf{ToG2.0}~\cite{Ma2024ThinkonGraph2D}} & \multicolumn{1}{c|}{\textbf{DoG}~\cite{ma2025debate}} & \multicolumn{1}{c|}{\textbf{GCR}~\cite{luo2024graph}} & \textbf{PDA}~\cite{Sun2024PyramidDrivenAP} &  \\ \hline
\multicolumn{1}{|c|}{\textbf{HotpotQA}} & \multicolumn{1}{c|}{41.0} & \multicolumn{1}{c|}{43.0} & \multicolumn{1}{c|}{42.6} & \multicolumn{1}{c|}{36.1} & \multicolumn{1}{c|}{43.0} & \multicolumn{1}{c|}{40.9} & \multicolumn{1}{c|}{45.3} & \multicolumn{1}{c|}{45.9} & 36.5 & \multicolumn{1}{c|}{\underline{60.0}} \\ \hline
\end{tabular}%
}
\caption{Comparison of existing RAG and GraphRAG methods with our proposed method on the Exact Match metric. On the TriviaQA dataset, the QA configuration with DeepSeek V3 and the combination of BeamSearch and WaterCircles algorithms (without restrictions on graph traversal) was used. On the HotpotQA dataset, the QA configuration with GPT-4o-mini and the same algorithm and graph restrictions was used.}
\label{tab:qamethods_compare}
\end{table}

As shown in Table \ref{tab:qamethods_compare}, our proposed method achieves a 14.1\% improvement over existing GraphRAG approaches in a specific configuration. However, it underperforms compared to standard RAG methods by 17.8\%. This discrepancy can be attributed to the fact that the RAG baselines evaluated in this study employed Reader and Retriever models that were specifically fine-tuned on the same dataset used for evaluation. As established in prior works, such in-domain fine-tuning typically yields optimal performance, and evaluation on out-of-domain datasets would be expected to result in significant degradation of measured metrics.

We also reproduce and evaluate the quality of HippoRAG method on DiaASQ and HotpotQA datasets: DiaASQ -- 0.53 (LLM-as-a-Judge); HotpotQA -- 60.2 (Exact Match). It can be seen that our method shows comparable results or outperforms HippoRAG.

\EOD

\end{document}